\documentclass[11pt]{article}

\usepackage[preprint]{acl}

\usepackage{times}
\usepackage{latexsym}

\usepackage[T1]{fontenc}

\usepackage{microtype}

\usepackage{inconsolata}

\usepackage{graphicx}
\usepackage{subcaption}

\usepackage{booktabs, tabularx}
\usepackage{xcolor, colortbl}
\usepackage{adjustbox}
\usepackage{float}
\definecolor{blue_box}{RGB}{228, 240, 249}
\definecolor{bbh}{RGB}{189,217, 243}
\definecolor{reddish}{RGB}{237,214, 206}

\usepackage{tcolorbox}
\tcbuselibrary{skins}
\tcbuselibrary{breakable}
\usepackage{titlesec}
\usepackage{multicol}
\usepackage{multirow}

\tcbset{bluetitle/.style={boxed title style={colback=white, top=1mm, bottom=-1mm, arc=1mm}}}

\newtcolorbox{promptbox}[2][]{
    enhanced,
    width=\textwidth,
    colback=blue_box,
    colframe=white,
    coltitle=black!50!white,
    boxrule=0pt,
    arc=1mm,
    fontupper=\ttfamily\tiny,
    fonttitle=\sffamily\small\bfseries,
    attach boxed title to top left={xshift=3mm,yshift=-2mm},
    title={#2},
    #1
}

\newtcolorbox{smallpromptbox}[2][]{
    enhanced,
    width=\linewidth,
    colback=blue_box,
    colframe=white,
    coltitle=black!50!white,
    boxrule=0pt,
    arc=1mm,
    fontupper=\ttfamily\tiny,
    fonttitle=\sffamily\small\bfseries,
    attach boxed title to top left={xshift=3mm,yshift=-2mm},
    title={#2},
    #1
}

\title{Are Language Models Script-Aware?}

 \author{David Kletz\footnotemark[2]\thanks{These authors contributed equally to this work.}, \hspace{0.8mm}   {\bf Sandra Mitrović\footnotemark[2]\footnotemark[1],} \hspace{0.8mm} {\bf  Ljiljana Dolamić\footnotemark[3],} \hspace{0.8mm} {\bf  Fabio Rinaldi\footnotemark[2]}\\
 \footnotemark[2] SUPSI, IDSIA, Switzerland\\
 \footnotemark[3] armasuisse, Science \& Technology, Switzerland\\
 {\tt \{david.kletz, sandra.mitrovic, fabio.rinaldi\}@supsi.ch} \\
 {\tt ljiljana.dolamic@armasuisse.ch}\\}

\begin{document}
\maketitle
\begin{abstract}

Language models frequently generate outputs in unintended languages or scripts, a phenomenon known as off-target generation. While existing research has focused on language selection, the dimension of script knowledge remains understudied: before any linguistic understanding can occur, users must recognize the graphic symbols in a model's response. We investigate whether Small and Large Language Models (SLMs and LLMs) possess script knowledge by testing them on multi-scriptic languages.
Through two complementary experiments, we evaluate whether models (1) adapt their output script to match the input, and (2) follow explicit instructions to generate text in a specified script.
The models we tested demonstrate substantial script knowledge: they all achieve a near-perfect Latin script fidelity (more than 98\%) and follow script instructions with high frequency.
Nevertheless, we notice differences between LLMs and SLMs, with higher scores for LLMs including for non-standard script combinations.

\end{abstract}

\section{Introduction}\label{s:intro}
Language models increasingly demonstrate remarkable multilingual capabilities, yet they frequently fail to generate outputs in the expected language, a phenomenon known as off-target generation \cite{xu2024a, richburg2024how}, also referred to as language confusion \cite{marchisio-etal-2024-understanding}. Several studies point the severity of this issue. For example, \citet{dai2024actmnmtautoconstrictionturningmultilingual} show that BLOOMZ-7B1 and mT0-xl \citep{muennighoff2022crosslingual} reach an off-target ratio higher than 60\% under a zero-shot setting in several languages. Likewise, Llama3.1 models generate non-English outputs in 37\% of cases when English is expected \cite{vykopal2025largelanguagemodelsmultilingual}. This problem is particularly frequent for low-resource languages \cite{sennrich-etal-2024-mitigating, nie2025mechanisticunderstandingmitigationlanguage}, further reinforcing the exclusion of speakers of low-resource languages.
While existing research has focused primarily on language selection, a more fundamental dimension remains understudied: script knowledge. Before any linguistic understanding can occur, users must recognize the graphic symbols employed in a model's response. Written communication relies on scripts--the sets of standardized visual symbols used to represent language \cite{MITON2021104771, gelb1963study}. This basic condition of legibility constitutes a prerequisite for meaningful human-AI interaction. Importantly, many languages can be written in multiple scripts (see details in Appendix~\ref{app:multi-mono_list}): Serbian uses both Latin and Cyrillic, Konkani is actively written in five scripts \cite{agarwal-etal-2025-script}, and Kazakh employs Latin, Cyrillic, and Arabic \citep{makhambetov-etal-2013-assembling}. These multi-scriptic languages provide natural test cases for assessing whether models possess genuine script knowledge and whether they exhibit systematic biases toward particular scripts.
Training corpora for language models are heavily biased toward the Latin script--more than 77\% of mBERT's vocabulary is Latin, compared to only 1\% Georgian \cite{acs2019bertvocab}. This imbalance measurably affects model behavior: the script used for a given language significantly impacts performance \cite{muller-etal-2021-unseen}, with, for example, dramatic output changes observed when Kazakh is written in Latin instead of Cyrillic \cite{isbarov2025tumluunifiednativelanguage}. While recent work has revealed performance disparities across scripts \cite{song2025mugevalproxyevaluationframework, hada-etal-2024-large}, linguistic gaps \cite{hussen2025statelargelanguagemodels}, and tokenization challenges \cite{zhang2025milicevalbenchmarkingmultilingualllms, land2025bpestaysscriptstructured}, as well as script tracing across transformer layers \cite{saji-etal-2025-romanlens,kletz2026scriptchoicellmsevidence}, no study has systematically evaluated whether models understand scripts as distinct entities that can be deliberately selected and controlled\footnote{As a preliminary step, we checked that SLMs can use different scripts when generating texts in a multi-scriptic language, details are in Appendix~\ref{app:script_fidelity}.}.
This work investigates the script knowledge of both Small and Large Language Models through the following research questions\footnote{Data available at \url{https://github.com/IDSIA-NLP/llm_script_awareness}}:
\begin{enumerate}
    \item[RQ1] Do models adapt their output script to match the input?
    \item[RQ2] Can models follow explicit instructions to generate text in a specified script, regardless of the input script?
\end{enumerate}

\section{Data and methods}\label{s:setup}

\subsection{Data}
\label{sec:data}
We restrict our analysis to languages that can be written in multiple scripts, along with mono-scriptic controls. Specifically, we examine:
Crimean Tatar (ISO 639-3: \texttt{crh}), a Turkic language written in both Latin and Cyrillic scripts, Kazakh (ISO 639-3: \texttt{kaz}), a Turkic language using Latin, Cyrillic and Arabic scripts, and Serbian (ISO 639-3: \texttt{srp}), a Slavic language using Latin and Cyrillic scripts.
We also add the following mono-scriptic languages as controls: French (ISO 639-3: \texttt{fra}) and
English (ISO 639-3: \texttt{eng}), both using Latin script only.

For each language, we select one dataset written in at least one dominant script of the language: \texttt{QIRIM}\footnote{\url{https://huggingface.co/datasets/QIRIM/crh-parallel-corpora}} for Crimean Tatar, providing aligned parallel texts in both Latin and Cyrillic scripts,
\texttt{arena-offline-qa}\footnote{\url{https://huggingface.co/datasets/kz-transformers/arena-offline-qa}} \citep{Sagyndyk2025KazRobertaConversational} for Kazakh in Cyrillic script,
\texttt{triviaqa} from the Serbian LLM evaluation suite\footnote{\url{https://huggingface.co/datasets/gordicaleksa/serbian-llm-eval-v1}} \citep{serbian-llm-eval} for Serbian in Cyrillic script,
\texttt{frenchQA}\footnote{\url{https://huggingface.co/datasets/CATIE-AQ/frenchQA}} \citep{frenchQA2023} for French in Latin script.

From each dataset, we randomly sample 1,000 questions of at least 100 characters. For QIRIM, we retain both Cyrillic and Latin versions of each text, yielding 2,000 prompts total.

\subsection{Models}
We evaluate both Small Language Models (SLMs) and Large Language Models (LLMs).
In particular, we evaluate the following Open-source SLM: Llama-2-7B-Chat \citep{touvron2023llama2openfoundation},  Mistral-7B-Instruct-v0.1 \citep{jiang2023mistral7b}, Mixtral-8x7B-Instruct-v0.1 \citep{jiang2024mixtralexperts}, Qwen1.5-7B-Chat \citep{bai2023qwentechnicalreport} and DeepSeek-Distill-Llama-8B \citep{deepseekai2025deepseekr1incentivizingreasoningcapability}.
We also evaluate the following API-based LLMs: GPT-4o-mini, GPT-4o \citep{openai2024gpt4ocard}, DeepSeek-Chat \citep{deepseekai2025deepseekv3technicalreport}, Claude-Haiku-3.5, Claude-Sonnet-4 \citep{anthropic2024claude}, Gemini-Flash-Lite, and Gemini-Flash \citep{comanici2025gemini25pushingfrontier}.

\subsection{Tokenization and Vocabulary Coverage}

A crucial question concerns the vocabulary size of models for each language-script pair. A potential issue we must address is whether certain words in a language-script pair are absent from the model's vocabulary (particularly in less frequent scripts). In such cases, models would be constrained to switch to scripts in which they can express their responses.

To assess whether vocabulary limitations could affect performance, we compute normalized vocabulary coverage scores for each model suite and language-script pair using representative corpora (for details, see Appendix~\ref{app:voc_size}).
Results show relatively consistent coverage (normalized scores 0.55–1.1), with language effects outweighing script effects. The exception is Kazakh-Arabic (score $\sim  0.6$ vs. $\sim 1.0$ for Latin), suggesting vocabulary limitations may partially explain lower performance for this particular configuration.

\subsection{Evaluation set-up}

Note that languages mentioned in Section \ref{sec:data} serve different roles in the study, depending on their specific properties and data availability; consequently, not every language is used to test every hypothesis.

\paragraph{Script Fidelity}
To assess whether models adapt their output script to match the input, we measure script fidelity as the percentage of cases where the output script matches the input script\footnote{For details, see Appendix~\ref{app:script_id}}. In this configuration, the predicted label is the detected output script, and the expected label is the input script. We also compile comprehensive statistics on all scripts produced by each model in response to different input scripts.

\paragraph{Script Forcing}

To evaluate whether models possess genuine script knowledge beyond mere input replication, we employ explicit script instructions. Models are prompted to generate responses in a specified script using the following template\footnote{All prompt variations are provided in Appendix~\ref{a:prompts}.}:
\begin{figure}[H]
\centering
\begin{smallpromptbox}[bluetitle]{English prompt}
Please answer the following question in \colorbox{bbh}{{language\_output}} using the \colorbox{bbh}{{script\_output}} script: \colorbox{bbh}{{question}}
\end{smallpromptbox}
\end{figure}
We compute the percentage of cases where the output script matches the requested script. This metric directly measures models' ability to follow explicit script instructions.

We test models' performance bounds using two reference points:

\begin{itemize}
    \item Upper bound: Natural language-script pairings (e.g., French-Latin, Serbian-Cyrillic) represent configurations where models should perform optimally.
    \item Lower bound: Artificial pairings (French-Cyrillic) measure models' capacity to follow instructions even for unnatural configurations.
\end{itemize}

Scores approaching the upper bound indicate the configuration is well-handled, while scores near the lower bound suggest the pairing is unnatural for the model.
Technical implementation details are provided in Appendix~\ref{app:tech_details}.

\section{Results}\label{s:results}

Our experiments reveal that language models possess substantial script knowledge, though with notable asymmetries. This knowledge manifests itself in two complementary ways: models can adapt their output script to match the input prompt, and they can follow explicit instructions to generate text in a specified script.

\subsection{Input Script Adaptation}

\begin{figure*}[!ht]
    \centering
    \subfloat{\includegraphics[width=0.47\textwidth]{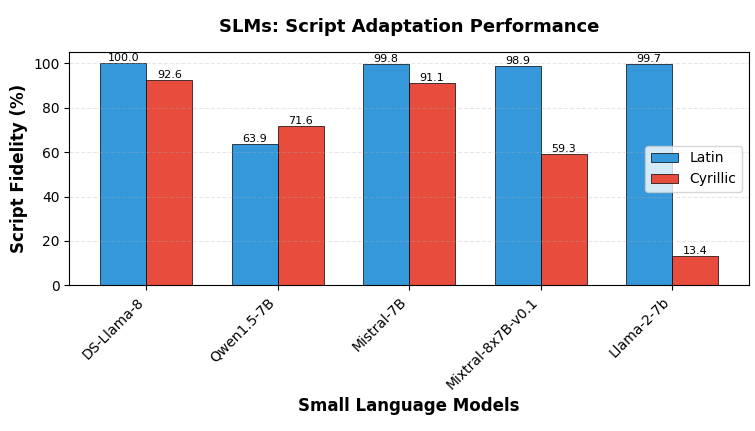}}
    \hspace{0.05\textwidth}
    \subfloat{\includegraphics[width=0.47\textwidth]{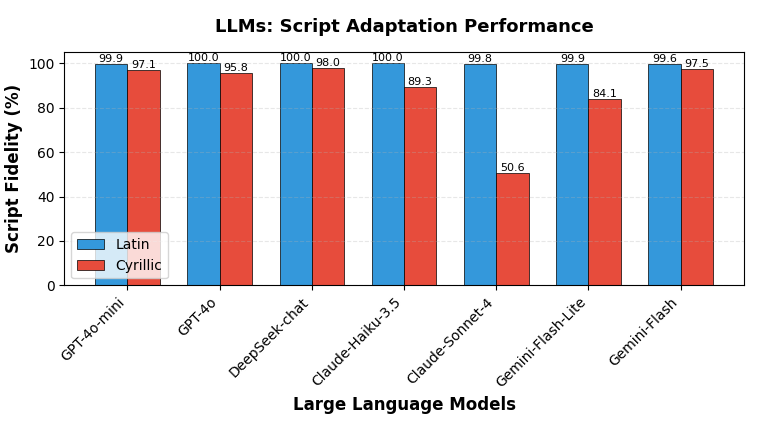}}

    \caption{Percentage of instances with matching input and output scripts for QIRIM dataset, per model.}
    \label{fig:adaptation_crimean_tatar}
\end{figure*}

When presented with prompts in different scripts, models demonstrate strong fidelity to the input script (see Figure~\ref{fig:adaptation_crimean_tatar}). For Latin prompts, all models achieve near-perfect alignment (lowest score: 98.9\%), maintaining consistency across both open-source SLMs and proprietary LLMs. However, Cyrillic prompts reveal a systematic bias: fidelity ranges from 13.4\% (Llama-2-7B) to 98\% (DeepSeek-chat), consistently lower than Latin performance. Notably, model size does not determine script fidelity: models as varied as DeepSeek-chat, Mistral-7B, and GPT-4o all achieve scores above 90\% for Cyrillic inputs.

However, the analysis of off-target scripts answers reveals differences between model types. API-based models respond almost exclusively in Latin or Cyrillic (with one exception: Gemini-Flash produced a single Greek output), while open-source SLMs display greater variability, sometimes generating outputs across eight different scripts when failing to match the input. This indicates that all models recognize the natural scripts for these languages, but open-source models exhibit less controlled script selection when deviating from the expected output.

\subsection{Required Script Adaptation}

\begin{figure*}[!h]
    \centering\includegraphics[width=\linewidth]{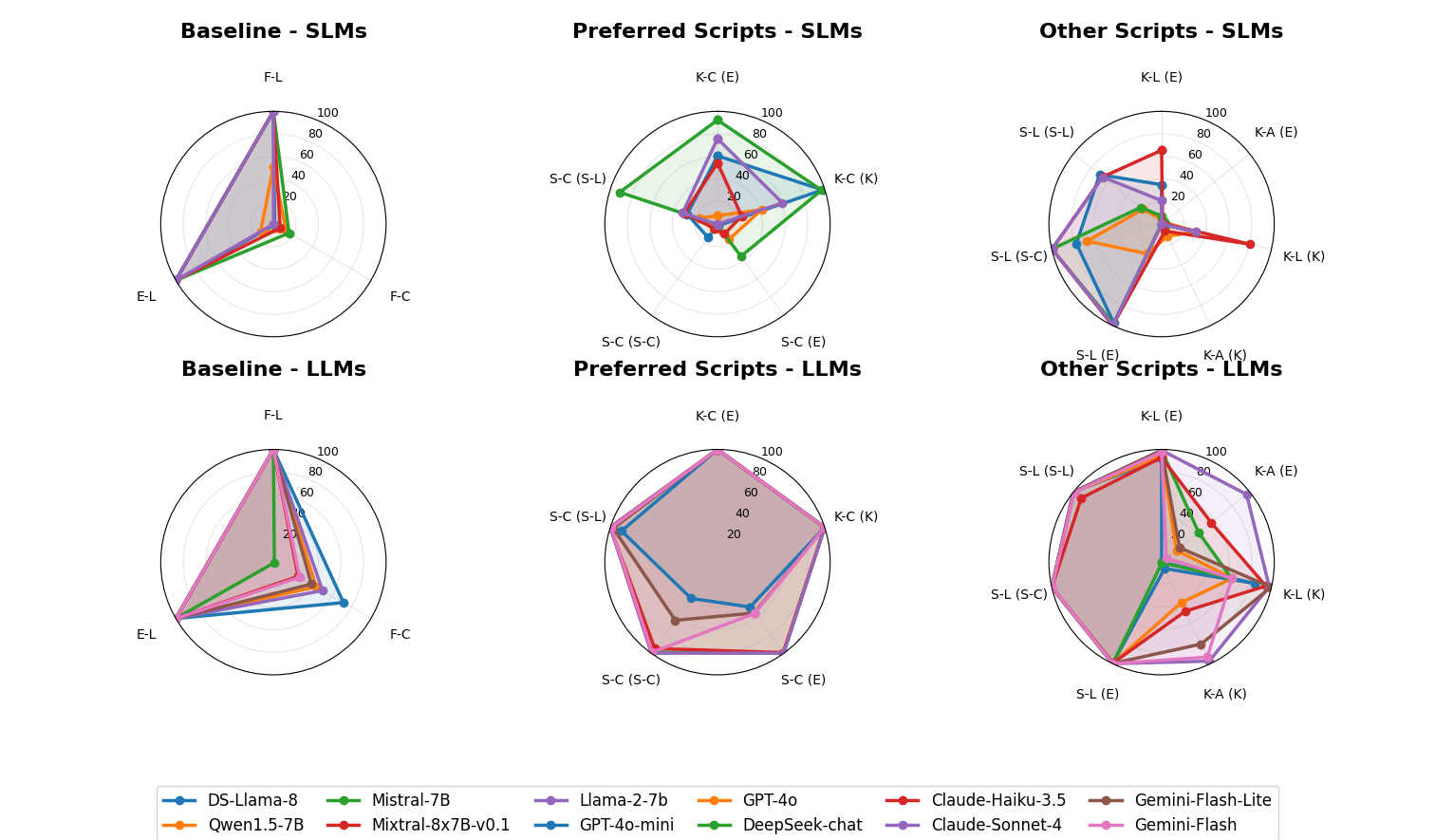}
    \caption{Script forcing results. Scores represent the percentage of cases where the script of the generated text matched the requested output script. Axes represent language-script pairs formatted as L-S, where L indicates the output language (S: Serbian, F: French, E: English, K: Kazakh) and S indicates the script (L: Latin, C: Cyrillic, A: Arabic). The language in parentheses indicates the prompt language. }
    \label{fig:results_forcing}
\end{figure*}

When explicitly instructed to use a particular script, models demonstrate robust script knowledge beyond simple imitation of the input script\footnote{As in previous experiments, Qwen1.5 exhibits substantially different behavior and is discussed separately in Appendix~\ref{app:qwen}.}. Performance varies systematically by language-script pairing: natural combinations like French-Latin, Kazakh-Cyrillic, and Serbian-Cyrillic consistently achieve the highest scores across most models (often reaching 100\%). In contrast, less common pairings such as Kazakh-Arabic or artificial combinations like French-Cyrillic show reduced performance, with the gap particularly pronounced for SLMs versus LLMs (see Figure~\ref{fig:results_forcing}).

Interestingly, some pairings contradict expectations (compared to the frequency of the pairs). Kazakh-Arabic often underperforms even the artificial French-Cyrillic combination (e.g., 0.0 vs. 71.8 for GPT-4o-mini), despite Arabic being a historically attested script for Kazakh. Similarly, Serbian-Latin consistently outperforms Serbian-Cyrillic, even though Cyrillic is the dominant script for Serbian, likely reflecting cross-lingual interference from Latin-script South Slavic languages like Croatian and Bosnian.

The substantial performance gap between LLMs and SLMs for non-natural script combinations (e.g., 71.8 vs. 16.0 for French-Cyrillic) stresses LLMs' high instruction-following capabilities. Additionally, prompt language introduces unexpected variance: changing only the prompt language while holding script and question language constant can shift performance by 10-12 points, suggesting that prompt language affects instruction interpretation beyond script knowledge itself.

\paragraph{Overall Findings}
These results demonstrate that current language models have acquired solid script knowledge. Multi-scriptic languages in our analysis receive outputs in more than one script, and models can both adapt to input scripts and follow explicit script instructions. However, this knowledge is not uniform: systematic biases toward certain scripts persist, natural language-script combinations are handled more reliably, and LLMs substantially outperform SLMs in flexible script manipulation, particularly for non-standard or artificial script requests.

\subsection{Quality of generated responses }

To verify that script changes do not systematically degrade response quality, we conducted an LLM-as-a-judge evaluation on a 10\% sample of generated responses across all configurations (see details in Appendix~\ref{app:quality}). Results show that factuality and completeness scores remain largely stable across script variations (typically within 0-10 points), even when script fidelity decreases. This confirms that models maintain semantic accuracy when generating text in non-preferred scripts, though compliance with script instructions may vary.

\section{Conclusions}

In this study, we have examined the extent to which language models possess awareness of scripts. Our findings reveal a substantial awareness that extends beyond simple pattern matching: Language Models can both adapt their output scripts to match input prompts and follow explicit instructions to generate text in specified scripts.

Models demonstrate consistently high Latin script fidelity but more variable performance with Cyrillic script. While all models recognize natural scripts for tested languages, natural language-script combinations are handled more reliably than non-standard pairings.
We also observed a performance gap between LLMs and SLMs, particularly for non-standard script combinations.

\section*{Limitations}

Our analyses in Section \ref{s:results} rely on GlotLID for detecting the script of generated text. The same tool is also used in the annotations of MultiQ. Although GlotLID achieves high accuracy, it is not completely reliable, and occasional misclassifications may influence our results.

Second, several of the languages examined are low- or mid-resource. The datasets used to train LLMs in these languages may themselves be biased, potentially leading to systematic biases in the models. For instance, while no precise statistics exist for the script distribution of Crimean Tatar, it is plausible that training data are skewed toward the Latin script, which could partly explain the models' observed preference for it over Cyrillic.

Finally, we consider only one artificial pairing (French-Cyrillic).

\section*{Acknowledgments}
The work described in this paper has been funded by the ``\textit{Language Preference and Hallucination Investigation in Multilingual RAG (LPHI-mRAG)}'' project, funded by armasuisse S\&T, Switzerland.

\bibliography{custom}

\appendix
\section{Vocabulary size per language}\label{app:voc_size}

\paragraph{Method} We retrieve a corpus for each language-script pair and tokenize it entirely using the model suite's tokenizer\footnote{For LLMs, we use token approximation tools (e.g., \url{https://huggingface.co/Xenova/claude-tokenizer}). We were unable to perform this analysis for Gemini models, which are therefore excluded from this experiment.}. We then count the number of types (unique tokens).
Since not all language-script pairs have corpora of equal size, and vocabulary size depends inversely on corpus size according to Zipf's law \citep{Zipf49}, we normalize our measurements (following Heaps' empirical law, \citet{heaps1978information}). We first calculate the size of each corpus used (available in Table~\ref{tab:summary_stats}).

\begin{table}[h]
\small
\centering
\begin{tabular}{lrrr}
\hline
Lang-script & Total words & Types & Sentences \\
\hline
Kz-ar & 27,805,939 & 810,230 & 57,827 \\
Kz-cyr & 4,656,620 & 301,614 & 371,902 \\
Kz-lat & 430 & 371 & 100 \\
Sr-cyr & 15,255 & 4,424 & 150 \\
Sr-lat & 569,206 & 89,537 & 2,919 \\
Cr-cyr & 517,595 & 109,847 & 77,688 \\
Cr-lat & 281,372 & 70,810 & 77,688 \\
Fr-lat & 8,073 & 2,484 & 800 \\
\hline
\end{tabular}
\caption{Summary statistics of datasets}
\label{tab:summary_stats}
\end{table}

Then, for each model suite and language-script pair, we compute a normalized vocabulary metric using the formula:
$$\frac{log(V)
}{log(N)}$$
where
$V$ is the vocabulary size (number of unique tokens) and $N$ is the corpus size (total tokens).

\begin{figure*}[!t]
    \centering
    \includegraphics[width=0.8\linewidth]{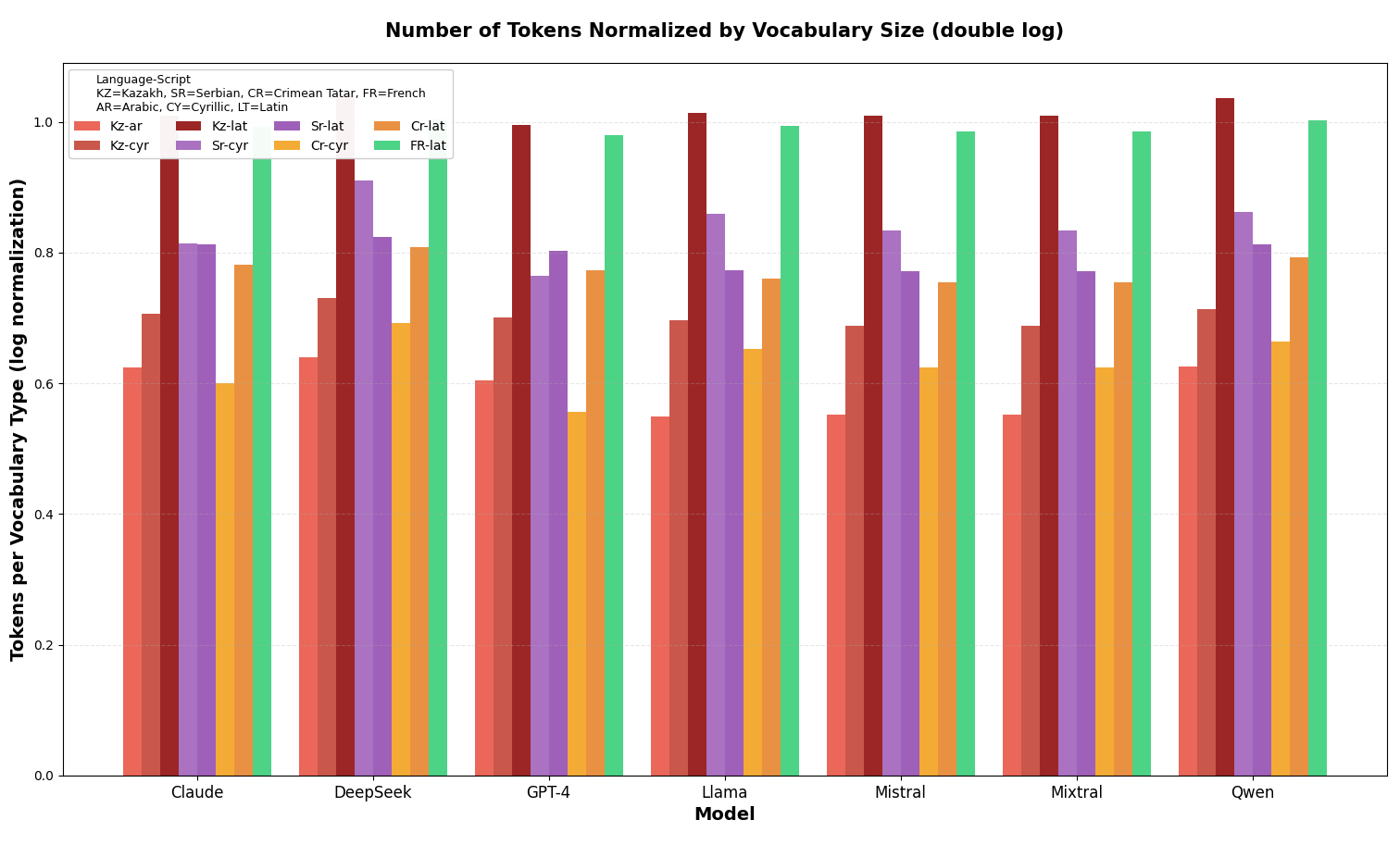}
    \caption{Size of the vocabulary (number of non-character tokens) per language-script combination, for each model. (with Zipf)}
\label{fig:size_vocab_lang_model}
\end{figure*}

\paragraph{Results} The results are displayed in Figure~\ref{fig:size_vocab_lang_model}. These findings demonstrate that models possess sufficient vocabulary coverage across language-script pairs to generate responses, even in less frequent scripts.

The results reveal a notable consistency in vocabulary sizes across model families. Normalized scores range between 0.55 and 1.1 for all model suites, indicating relatively balanced coverage.
The most significant effect is attributable to language rather than script. For instance, Serbian scores consistently fall between 0.75 and 0.9, while French scores are between 0.95 and 1.05. This variation likely reflects differences in morphological complexity and corpus characteristics rather than fundamental vocabulary gaps.

Within individual languages, the choice of script produces only modest variation in vocabulary coverage. Latin-script pairs consistently achieve slightly higher scores, but this advantage is small and does not uniformly indicate preferential treatment--the Latin script is not always dominant. The notable exception is Kazakh: Latin script exhibits substantially larger vocabulary coverage (normalized score $\sim 1.0$) compared to Arabic script ($\sim 0.6$). This disparity should be considered when interpreting results.

\section{Quality control}\label{app:quality}

\subsection{Responses Quality}
We verify whether response quality is affected by the required output script. Specifically, we examine whether switching to a non-primary script for a given language results in substantial degradation of response quality.

\paragraph{Evaluation methodology} We assess response quality using an LLM-as-a-judge approach \cite{NEURIPS2023_91f18a12}. For each model and configuration (see Section~\ref{s:setup}), we randomly sample 10\% of the generated responses and pair them with their corresponding gold answers from the original datasets. These pairs are then provided to a judge LLM, which evaluates each response according to two criteria: (1) Factuality: Is the generated answer factually accurate and semantically aligned with the gold answer? (2) Completeness: Does the generated answer cover all essential information, regardless of the script used? Each criterion is scored on a 3-point scale (0, 1, or 2).
To avoid bias, we use \textit{gpt-5-nano} as the judge model, which was not included in our main experiments. We compute the average score for each configuration and model.

\paragraph{Comparative analysis} Since our objective is to assess quality changes across scripts, we calculate score differences between the following configuration pairs:

\begin{itemize}
    \item French-Latin vs. French-Cyrillic
    \item Serbian-Cyrillic vs. Serbian-Latin (prompt in Serbian-Latin)
    \item Serbian-Cyrillic vs. Serbian-Latin (prompt in Serbian-Cyrillic)
    \item Serbian-Cyrillic vs. Serbian-Latin (prompt in English-Latin)
    \item Kazakh-Cyrillic vs. Kazakh-Arabic (prompt in Kazakh-Cyrillic)
    \item Kazakh-Cyrillic vs. Kazakh-Latin (prompt in Kazakh-Cyrillic)
    \item Kazakh-Cyrillic vs. Kazakh-Arabic (prompt in English-Latin)
    \item Kazakh-Cyrillic vs. Kazakh-Latin (prompt in English-Latin)
\end{itemize}

For each pair, we compute the score difference by subtracting the second configuration's score from the first (preferential) configuration's score. Positive differences indicate that the preferential script configuration yielded higher quality outputs, while negative differences indicate the opposite. If models possess genuine script-independence, these score differences should remain low across all comparisons.

Results are presented in Tables~\ref{tab:quality_differences_SLMs} and \ref{tab:quality_differences_LLMs}.

\begin{table*}[t]
\centering
\tiny
\setlength{\tabcolsep}{5pt}
\begin{tabular}{l|ccc|ccc|ccc|ccc|ccc|ccc|ccc}
\toprule
\multirow{2}{*}{\textbf{Config}} & \multicolumn{3}{c|}{\textbf{gpt-4o-m}} & \multicolumn{3}{c|}{\textbf{gpt-4o}} & \multicolumn{3}{c|}{\textbf{Deepseek-chat}} & \multicolumn{3}{c|}{\textbf{Haiku}} & \multicolumn{3}{c|}{\textbf{Sonnet}} & \multicolumn{3}{c}{\textbf{Gemini-flash}} & \multicolumn{3}{c}{\textbf{Gemini}} \\
& F & C & S & F & C & S & F & C & S & F & C & S & F & C & S & F & C & S & F & C & S \\
\midrule
F-C vs F-L & -4 & 0 & -28 &-8 & -4 & -56 &-16 & -14 & -99 &-17 & -18 & -75 &-18 & -17 & -50 &-7 & -4 & -73 &-13 & -16 & -61 \\
S-C vs S-L (S-L) & -3 & -9 & -10 &-2 & -5 & 0 &-4 & -4 & 0 &-2 & -8 & 9 &-6 & -8 & 0 &-12 & -15 & 0 &0 & -6 & -3  \\
S-C vs S-L (S-C) & -8 & -11 & -60 &-6 & -9 & 0 &-4 & -5 & -1 &6 & 1 & -5 &-4 & -6 & 0 &-6 & -10 & -1 &-1 & -2 & -36 \\
S-C vs S-L (E) & 6 & 5 & -51 &2 & -1 & -1 &0 & 4 & 0 &-9 & -10 & 0 &-3 & -4 & 0 &-24 & -29 & -44 &-10 & -8 & -44 \\
K-C vs K-A (K) & -21 & -13 & -8 &-15 & -9 & -6 &4 & 5 & -1 &-9 & 7 & -7 &-2 & 4 & -1 &10 & 9 & -2 &-5 & 1 & -1 \\
K-C vs K-L (K) & 2 & 0 & -100 &-7 & 2 & -83 &-4 & 0 & -58 &-38 & -7 & -44 &-20 & -2 & -4 &-2 & 2 & -95 &-6 & 0 & -80 \\
K-C vs K-A (E) & -16 & -13 & -15 &-6 & -4 & -36 &-4 & 0 & -36 &-14 & 2 & -5 &1 & 4 & -2 &-1 & 3 & -36 &-5 & 0 & 0 \\
K-C vs K-L (E) & 2 & 2 & -94 &-18 & -6 & -60 &-8 & -2 & -99 &-34 & -7 & -52 &-11 & -1 & -2 &-3 & 6 & -6 &-11 & 0 & -19 \\
\bottomrule
\end{tabular}
\caption{Quality score differences across script configurations of LLMs. F: Factuality, C: Completeness, S: Script fidelity. Lines name are language-script pairs formatted as L-S, where L indicates the output language (S: Serbian, F: French, E: English, K: Kazakh) and S indicates the script (L: Latin, C: Cyrillic, A: Arabic). The language in parentheses indicates the prompt language. }
\label{tab:quality_differences_SLMs}
\end{table*}

\begin{table*}[t]
\centering
\tiny
\begin{tabular}{l|ccc|ccc|ccc|ccc}
\toprule
\multirow{2}{*}{\textbf{Config}} & \multicolumn{3}{c|}{\textbf{DeepSeek-R1-Distill-Llama-8B}} & \multicolumn{3}{c|}{\textbf{Llama-2-7B-Chat}} & \multicolumn{3}{c|}{\textbf{Mistral-7B-Instruct}} & \multicolumn{3}{c}{\textbf{Mixtral-8x7B-Instruct}} \\
& F & C & S & F & C & S & F & C & S & F & C & S  \\
\midrule
F-C vs F-L & 5 & 2 & -99 &-1 & -2 & -100 &-4 & 2 & -84 &-4 & -7 & -93 \\
S-C vs S-L (S-L) & 0 & -4 & -42 &-5 & -3 & -34 &2 & 0 & 68 &-2 & -3 & -36\\
S-C vs S-L (S-C) & 2 & 2 & -63 &-4 & -2 & -99 &-11 & -10 & -99 &4 & 8 & -95 \\
S-C vs S-L (E) &1 & 4 & -96 &-1 & 1 & -99 &-11 & -10 & -63 &2 & -2 & -89 \\
K-C vs K-A (K) & 3 & 6 & -26 &0 & 0 & -55 &0 & 0 & -85 &10 & 7 & 11\\
K-C vs K-L (K) & 0 & 2 & -60 &0 & 0 & -76 &-1 & 0 & -90 &-1 & 0 & -53 \\
K-C vs K-A (E) & -1 & 0 & -98 &2 & 0 & -30 &0 & 1 & -93 &6 & 0 & 58  \\
K-C vs K-L (E) & 0 & 0 & -98 &3 & 1 & -61 &0 & 0 & -97 &0 & 1 & -16\\
\bottomrule
\end{tabular}
\caption{Quality score differences across script configurations of SLMs. F: Factuality, C: Completeness, S: Script fidelity. Lines name are language-script pairs formatted as L-S, where L indicates the output language (S: Serbian, F: French, E: English, K: Kazakh) and S indicates the script (L: Latin, C: Cyrillic, A: Arabic). The language in parentheses indicates the prompt language.}
\label{tab:quality_differences_LLMs}
\end{table*}

\paragraph{Results} The results reveal a general trend of quality preservation across script changes: the majority of score variations (for both completeness and factuality) range between 0 and -10 points.
When examined alongside the decrease in script fidelity rates, these findings demonstrate that script changes do not necessarily lead to diminished response quality. This suggests that models can maintain semantic accuracy even when generating text in non-preferred scripts--though they may struggle to comply with the script instruction itself.

\paragraph{Model-specific patterns} SLMs exhibit the smallest quality variations (between -7 and +10 points), while LLMs show larger fluctuations. This difference may stem from SLMs' more limited knowledge base, resulting in lower baseline scores even with languages' primary scripts, thereby leaving less room for degradation.
Among LLMs, behavior varies substantially by model family. While GPT models show moderate score variations, Claude models exhibit considerably larger differences, particularly for configurations with Kazakh.

\paragraph{Metric consistency} The two evaluation criteria show strong agreement: factuality and completeness scores consistently exhibit similar signs and magnitudes.

\subsection{Perplexity Analysis}

Our second analysis examines model confidence in text generation across different scripts through perplexity measurements.

\paragraph{Methodology} For each SLM\footnote{Since LLMs are accessed via API, we do not have access to token probabilities and cannot compute perplexity.}, we calculate perplexity scores over each complete generated response. The objective is to compare, for each model and language, the perplexity when using the primary script versus alternative scripts.

\begin{figure*}[t]
    \centering

    \begin{subfigure}{0.45\textwidth}
        \centering
        \includegraphics[width=0.95\textwidth]{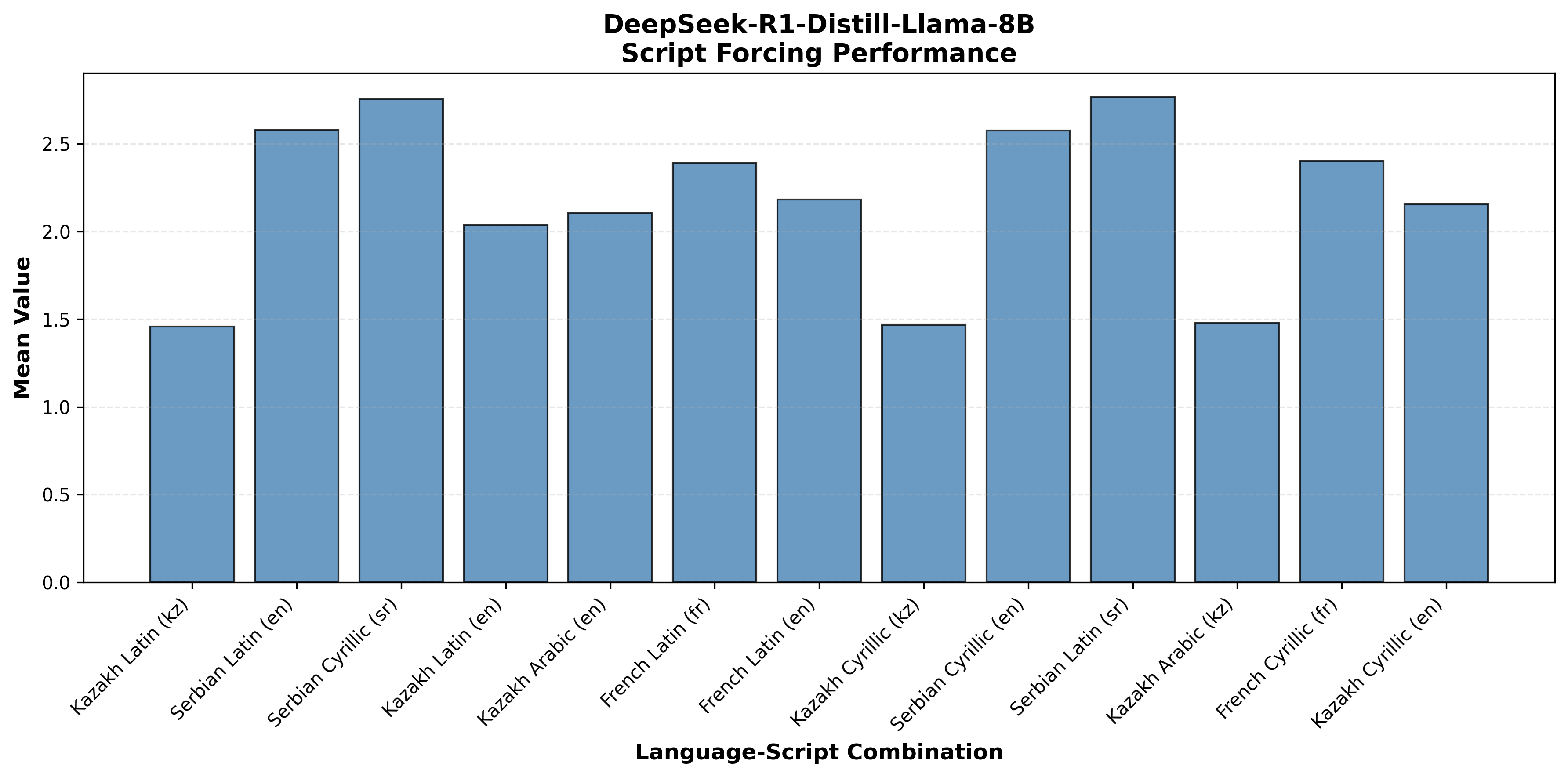}
        \caption{DeepSeek-R1-Distill-Llama-8B}
        \label{fig:perplexity_deepseek}
    \end{subfigure}
    \hfill
    \begin{subfigure}{0.45\textwidth}
        \centering
        \includegraphics[width=0.95\textwidth]{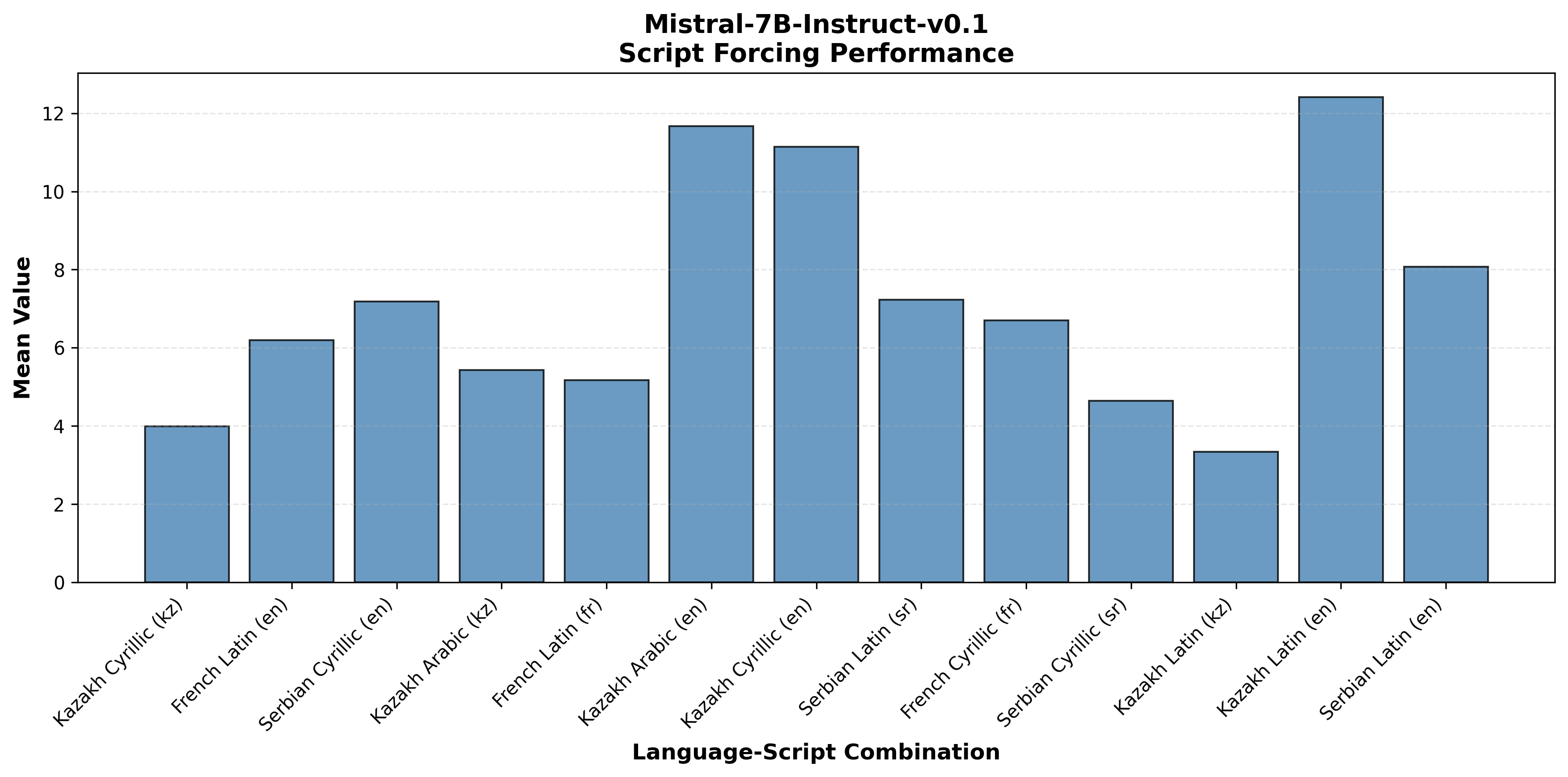}
        \caption{Mistral-7B-Instruct-v0.1}
        \label{fig:perplexity_mistral}
    \end{subfigure}

    \vspace{0.3cm}

    \begin{subfigure}{0.45\textwidth}
        \centering
        \includegraphics[width=0.95\textwidth]{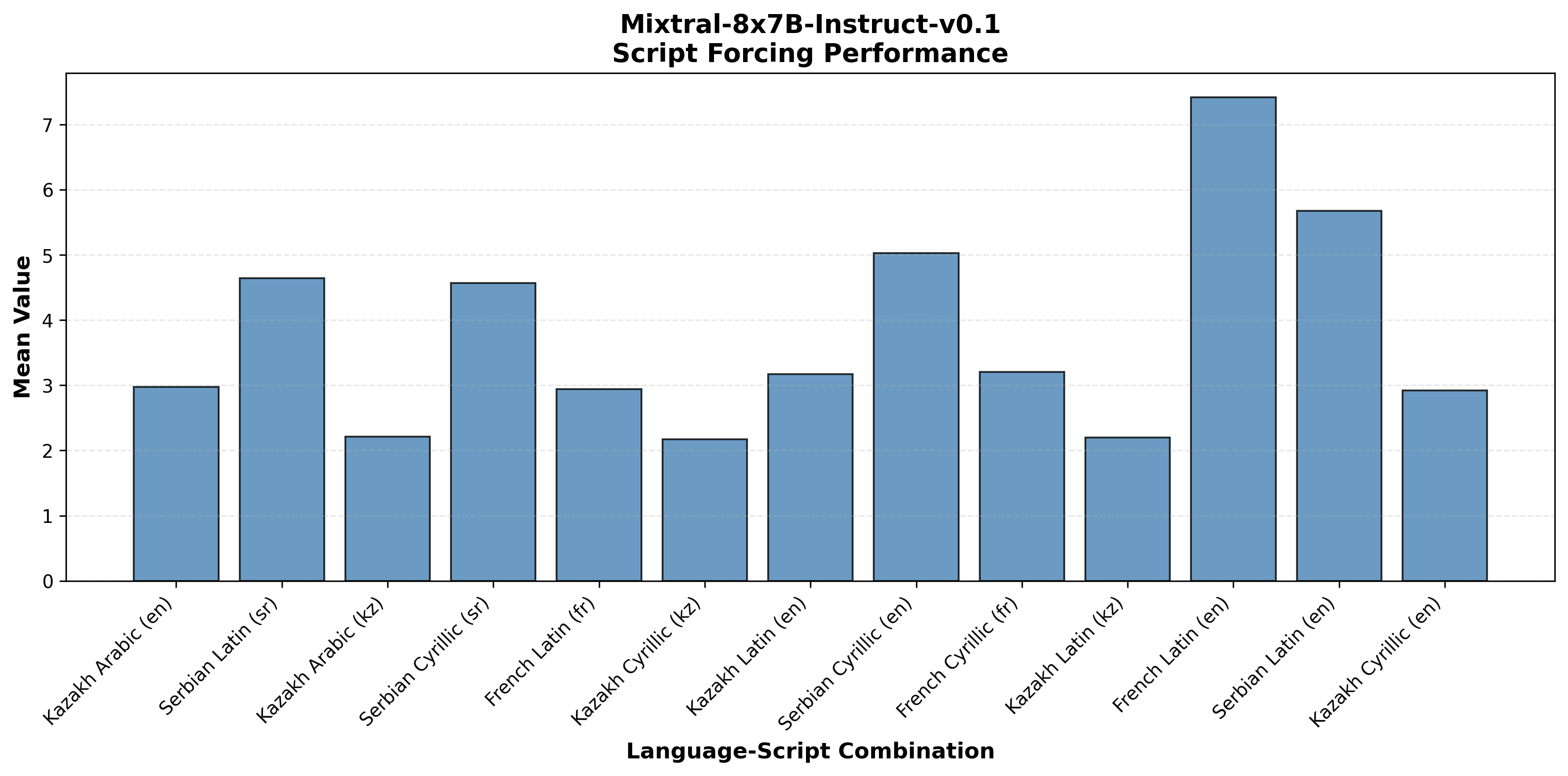}
        \caption{Mixtral-8x7B-Instruct-v0.1}
        \label{fig:perplexity_mixtral}
    \end{subfigure}
    \hfill
    \begin{subfigure}{0.45\textwidth}
        \centering
        \includegraphics[width=0.95\textwidth]{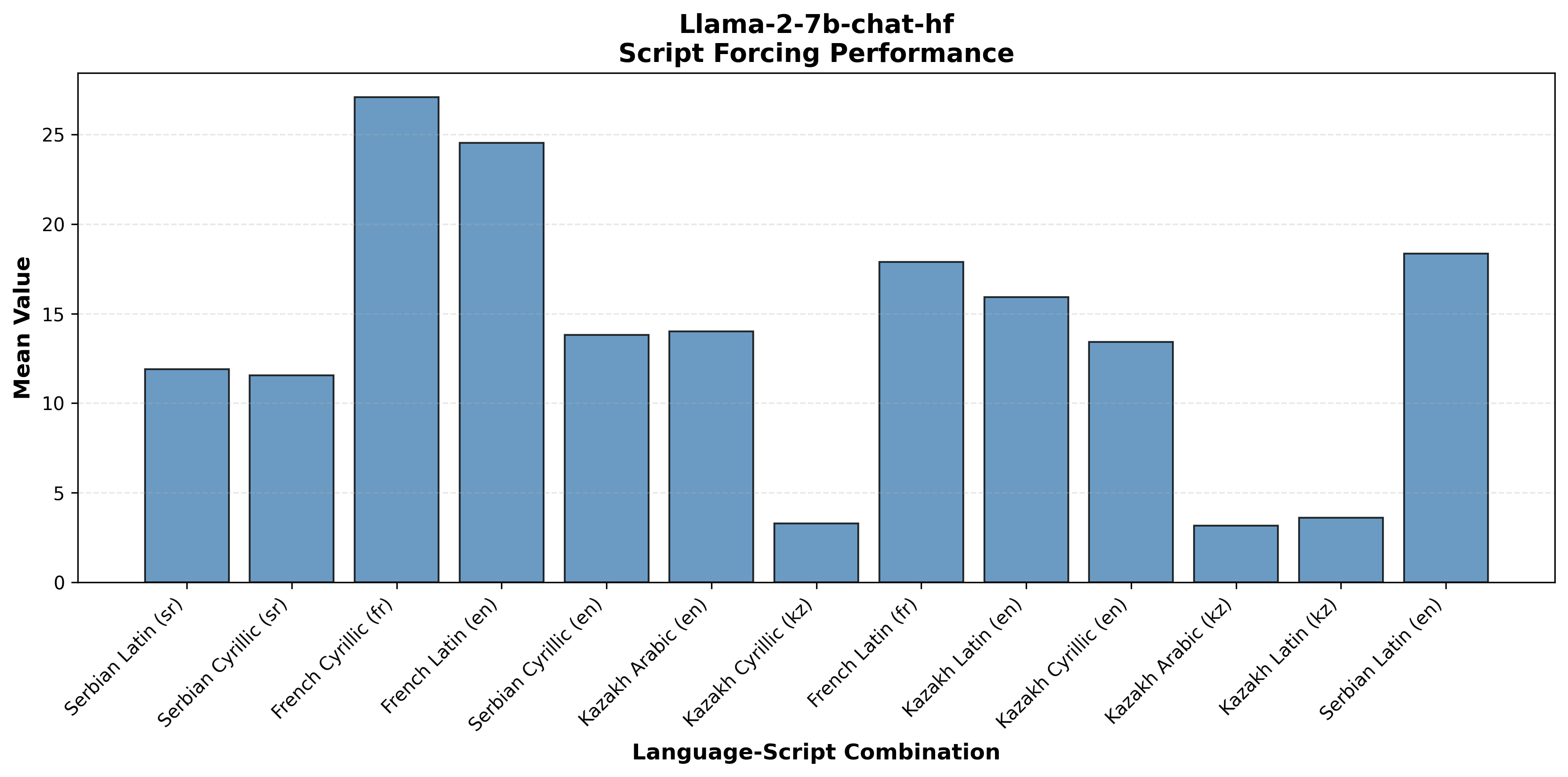}
        \caption{Llama-2-7B-Chat-HF}
        \label{fig:perplexity_llama}
    \end{subfigure}

    \vspace{0.3cm}

    \begin{subfigure}{0.45\textwidth}
        \centering
        \includegraphics[width=0.95\textwidth]{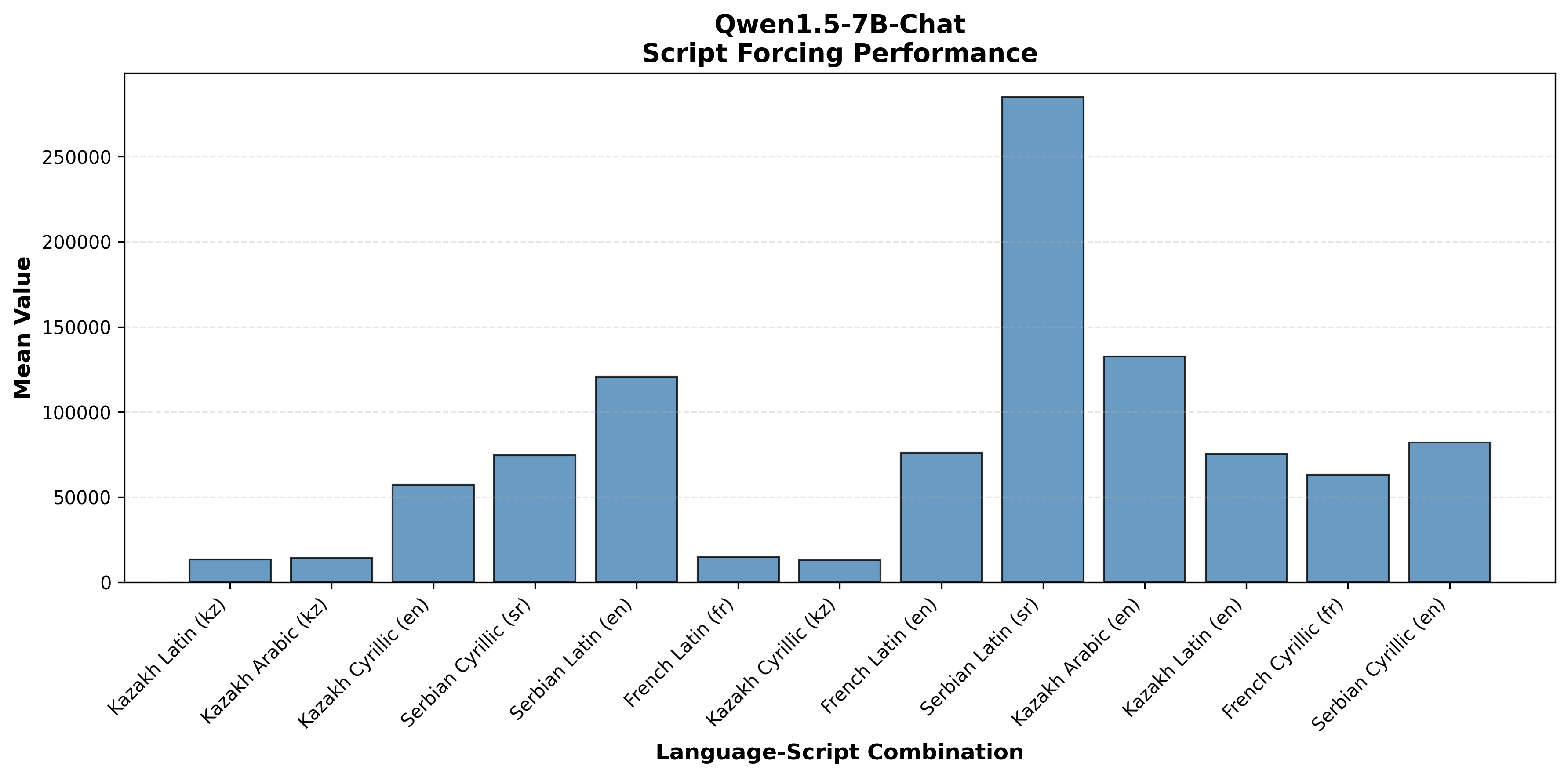}
        \caption{Qwen1.5-7B-Chat}
        \label{fig:perplexity_qwen}
    \end{subfigure}

    \caption{Perplexity distributions across script configurations for each SLM. Each panel shows perplexity values for different language-script combinations, with the language of the prompt in parenthesis.}
    \label{fig:perplexity_all}
\end{figure*}

Results are presented in Figures~\ref{fig:perplexity_all}.

With the notable exception of Qwen1.5-7B (discussed in Appendix~\ref{app:qwen}), models exhibit relatively stable perplexity across scripts, confirming their confidence in generating the target language regardless of the writing system used.

Several noteworthy patterns emerge from the analysis. First, Kazakh-Arabic does not yield the highest perplexity, despite being the least common script for this language. This could be explained by low script fidelity in this configuration: when instructed to use Arabic script, models often default to other scripts. Second, DeepSeek-R1-Distill-Llama demonstrates the most stable perplexity across all configurations, suggesting robust cross-script generation capabilities.

The script used in the prompt itself significantly affects generation confidence. For instance, Mistral-7B exhibits substantially higher perplexity when prompted in Serbian-Cyrillic and asked to respond in Serbian-Latin compared to when asked to respond in Serbian-Cyrillic (and vice versa for Latin-to-Cyrillic transitions). This asymmetry suggests that aligning the prompt script with the expected output script reduces model uncertainty, even when both scripts are natural for the language. This finding indicates a practical recommendation: when requesting responses in a specific script, using that same script in the prompt yields more confident generations than relying solely on the language's dominant script.

\section{Multi- and Mono-scriptic languages}\label{app:multi-mono_list}

An overview of literature sources mentioning certain languages as multi-scriptic can be seen in Table~\ref{tab:appendix_multi_literature}.

Given there is no comprehensive existing categorization of languages on mono-scriptic and multi-scriptic, we cross-check Tatoeba and literature sources (Table~\ref{tab:appendix_multi_literature}). Tatoeba consist of 166 different languages but we were able to retrieve ISO codes for only 97 of these. Only 12 out of these 97 languages are considered as multi-scriptic in Tatoeba. Table~\ref{tab:num_lang_per_script_cat} presents number of languages per category (``mixed'', ``multi'', ``mono'') according to both Tatoeba and literature.
The list of ``mixed'' languages (where Tatoeba and literature disagrees in the number of scripts used per language) as well as the list of multi-scriptic languages (where both sources agree that multiple scripts can be used per language) are displayed in Table~\ref{tab:mixed_langs_list}.

\begin{table}[!ht]
\centering
\setlength{\tabcolsep}{4pt}
\begin{tabularx}{\linewidth}{p{3.6cm} | p{3.7cm} }
\toprule
\textbf{Language Script Type} & \textbf{Number of Languages} \\
\midrule
Mono-scriptic    &  78 \\
Multi-scriptic   &  10 \\
Mixed   &   9 \\
\bottomrule
\end{tabularx}
\caption{Number of languages in different script-based categories based on Tatoeba and literature. Mono-scriptic (multi-scriptic) are those languages for which both sources agree that they use a single (multiple) script(s), respectively. Mixed category languages use different number of scripts (mono in Tatoeba and multi in the literature, or vice versa).}
\label{tab:num_lang_per_script_cat}
\end{table}

\begin{table}[!ht]
\centering
\setlength{\tabcolsep}{4pt}
\begin{tabularx}{\linewidth}{p{3.4cm} p{3.5cm}}
\toprule
\textbf{Mixed Category} & \textbf{Multi-script} \\
 & \textbf{Category} \\
\midrule
 Bulgarian & Bosnian\\
 Kurdish (Central) & Chavacano \\
 Javanese & Kunming \\
 Korean & Kazakh \\
 Marathi & Kurmanji \\
 Macedonian & Ladino \\
 Tagalog & Turkmen \\
 Uyghur & Cantonese \\
 Ukrainian & Malay (Kuala Lumpur) \\
  & Malay \\
\bottomrule
\end{tabularx}
\caption{\textbf{Left:} Mixed category languages according to Tatoeba and literature (using different number of scripts in Tatoeba with respect to the literature). \textbf{Right:} Multi-script category containing languages which both Tatoeba and literature consider multi-scriptic.}
\label{tab:mixed_langs_list}
\end{table}

\begin{table*}[!ht]
\centering
\setlength{\tabcolsep}{4pt}
\begin{tabularx}{\linewidth}{p{11cm} | p{3cm} }
\toprule
\textbf{Source URL} & \textbf{Language} \\
\midrule
\multirow{9}{*}{https://sites.google.com/site/seesscm/languages-in-multiple-scripts} & Bosnian\\
  & Bulgarian\\
  & Kurdish (Central)\\
  & Macedonian\\
  & Moldavian \\
  & Tajik\\
  & Turkmen\\
  & Ukrainian\\
   & Tatar\\
\midrule
\multirow{5}{*}{https://en.wikipedia.org/wiki/Digraphia} & Javanese\\
  & Kazakh\\
  & Konkani\\
   & Marathi\\
   & Malay\\
\midrule
\multirow{2}{*}{https://journals.sagepub.com/doi/epub/10.1177/18793665231188380} & Serbian-Croatian\\
 & Uzbek (Northern)\\
\midrule
https://www.ibidem.eu/out/media/vlb\_9783838217932\_2.pdf & Korean\\
\bottomrule
\end{tabularx}
\caption{Literature sources (left) explicitly mentioning multi-scriptic feature of a language (right).}
\label{tab:appendix_multi_literature}
\end{table*}

\section{Scripts used by SLM}\label{app:script_fidelity}

As explained in Section~\ref{s:intro}, all languages are associated with scripts. Some languages are mono-scriptic, written consistently in a single script, while others are multi-scriptic, allowing multiple coexisting scripts.
As a preliminary step, we investigate whether language models are constrained to a fixed set of scripts when generating text for a given language, or whether they can produce outputs in multiple scripts. In particular, we test whether models use only one script for mono-scriptic languages, and can exploit several scripts for multi-scriptic ones.

\subsection{Mono- vs. Multi-script Languages}\label{ss:multiVSmono}

Determining whether a language is mono- or multi-scriptic is not straightforward: the existence of a corpus written in a given script does not imply that this script is the only one used for that language.
To the best of our knowledge, no exhaustive classification of world languages by their script multiplicity currently exists. We therefore rely on two complementary resources to establish our ground truth.

We first extract information from the Tatoeba Translation Challenge \citep{tiedemann-2020-tatoeba}\footnote{Data from the September 2023 edition, available at \url{https://github.com/Helsinki-NLP/Tatoeba-Challenge/tree/master}}.
Its metadata includes, for a number of languages, explicit annotations of the script used--particularly for cases where multiple scripts are possible.
Languages for which Tatoeba specifies a script are treated as ``multi-scriptic according to Tatoeba''; those without such annotations are considered ``mono-scriptic according to Tatoeba''.

We complement these data with scholarly sources identifying specific multi-scriptic languages. All languages mentioned in the literature as employing multiple scripts are labeled ``multi-scriptic according to the literature''.
Because these works focus only on positive cases, they do not allow us to infer that unlisted languages are mono-scriptic.

Combining the two sources sometimes yields discrepancies: some languages appear as mono-scriptic in Tatoeba yet multi-scriptic in the literature. We therefore define three categories:
(1) Mono – agreement between Tatoeba and the literature that the language uses one script;
(2) Multi – agreement that it uses several;
(3) Mixed – disagreement between the two sources.
The detailed composition of these groups is provided in Appendix~\ref{app:multi-mono_list}.

\subsection{Experimental Setup}
For each language we compare the number of distinct scripts used by the models in generation with the corresponding ground-truth number defined in \ref{ss:multiVSmono}.

\paragraph{Method}
We employ an open-question answering scenario: questions are posed in various languages and scripts, and models are allowed to respond freely--without conditioning on a target language or script.
For each language, we identify the scripts present in the generated outputs and record whether a single or multiple scripts were used.
We then compute the proportion of languages whose outputs include more than one script, broken down by the Mono, Multi, and Mixed categories.

\paragraph{Data}
We rely on the MultiQ dataset \citep{holtermann-etal-2024-evaluating}, which contains 200 questions across 137 input languages, along with model responses. MultiQ is particularly suitable for this study as it provides multilingual outputs without explicit script constraints.
In total, 29 distinct output scripts are observed in the dataset.
Because this section aims to provide a preliminary exploration motivating subsequent experiments, we directly reuse the MultiQ data without additional fine-tuning or resampling.

\paragraph{Models}
Following the scope of MultiQ, we focus on six open-source SLMs:
Llama-2-(7/13/70)B-Chat-HF \citep{touvron2023llama2openfoundation}, Mistral-7B-Instruct-v0.1 \citep{jiang2023mistral7b}, Mixtral-8×7B-Instruct-v0.1 \citep{jiang2024mixtralexperts}, and Qwen 1.5-7B-Chat \citep{bai2023qwentechnicalreport}.
This selection allows us to observe general tendencies before extending the analysis to a broader set of larger models.

\subsection{Results}

\begin{figure*}[!ht]
    \centering
    \subfloat[Llama-2-(7b/13b/70b)-chat-hf]{\includegraphics[width=0.4\textwidth]{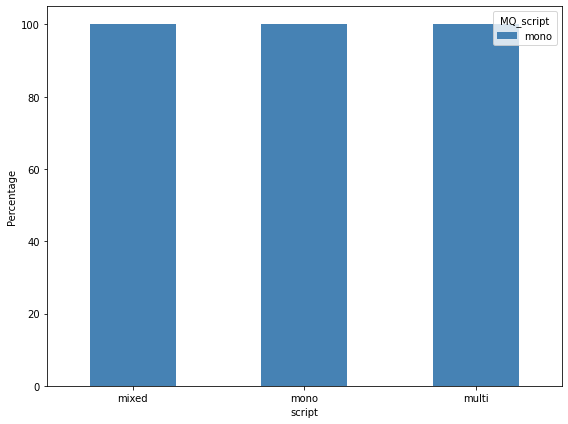}}
    \subfloat[Mistral-7B-Instruct-v0.1]{\includegraphics[width=0.4\textwidth]{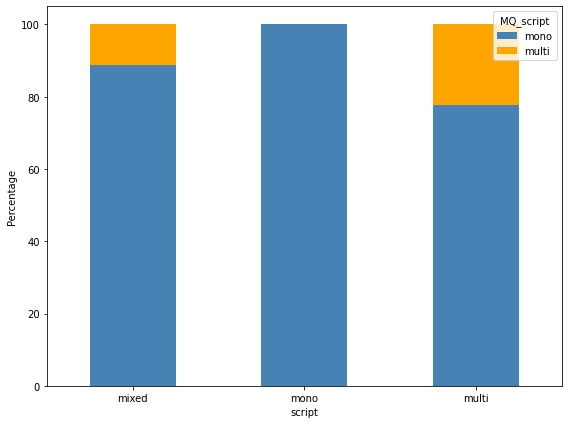}}
    \\
    \subfloat[Mixtral-8x7B-Instruct-v0.1]{\includegraphics[width=0.4\textwidth]{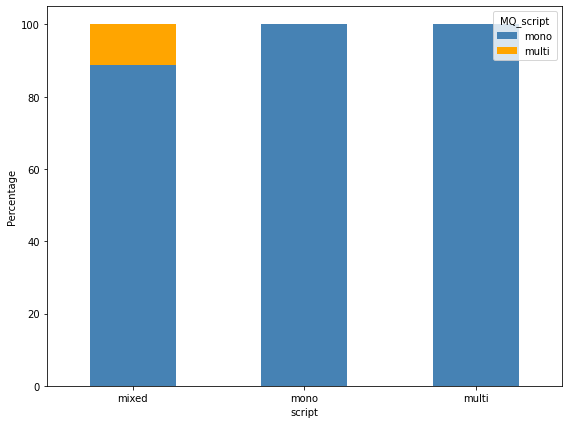}}
    \subfloat[Qwen1.5-7B-Chat]{\includegraphics[width=0.4\textwidth]{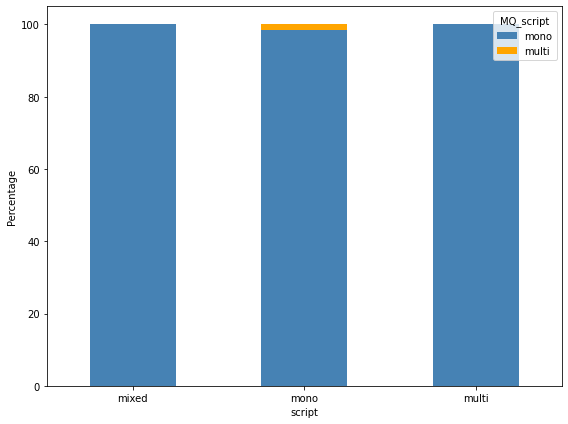}}
    \caption{Percentage of languages by script usage. Blue bars represent languages for which the model generates text in a single script, while orange bars represent languages for which the model uses two or more scripts. The input groups correspond to the language categories defined in Sub-Section~\ref{ss:multiVSmono}.}
    \label{fig:mono-multi}
\end{figure*}

Figure~\ref{fig:mono-multi} summarizes, for each model, the proportion of languages generated in one or multiple scripts across the three language categories (mono, multi, mixed).

Three out of the four models exhibit non-zero proportions of multi-script generations (orange bars). This observation indicates that these models can, at least in some cases, generate outputs using more than one script for the same language. Such behavior suggests that the models have acquired at least a partial knowledge of script multiplicity.

Nonetheless, mono-script generations (blue bars) largely dominate the results. The use of multiple scripts remains marginal, even within the multi-script language group. In three plots, all multi-script languages are generated exclusively in a single script. Only Mistral-7B-Instruct-v0.1 displays a notable deviation from this pattern, producing multi-script outputs for approximately 30\% of the multi-script languages.
An important consideration is that each entry in MultiQ is written in a single input script. The apparent predominance of mono-script outputs may therefore reflect an alignment between the input and output scripts, rather than a genuine lack of multi-script competence. This hypothesis is investigated further in Section~\ref{s:results}.

Languages classified as ``mixed'' exhibit variable behavior across models. Two models generate outputs in multiple scripts for some of these languages, while the remaining two restrict their generations to a single script.

\paragraph{}
Overall, these results confirm that several models are capable of producing text in more than one script for a given language.

\section{Infidelity details}\label{app:infidelity_details}

\begin{figure*}[!h]
    \centering
    \begin{subfigure}{0.45\textwidth}
        \centering
        \includegraphics[width=\linewidth]{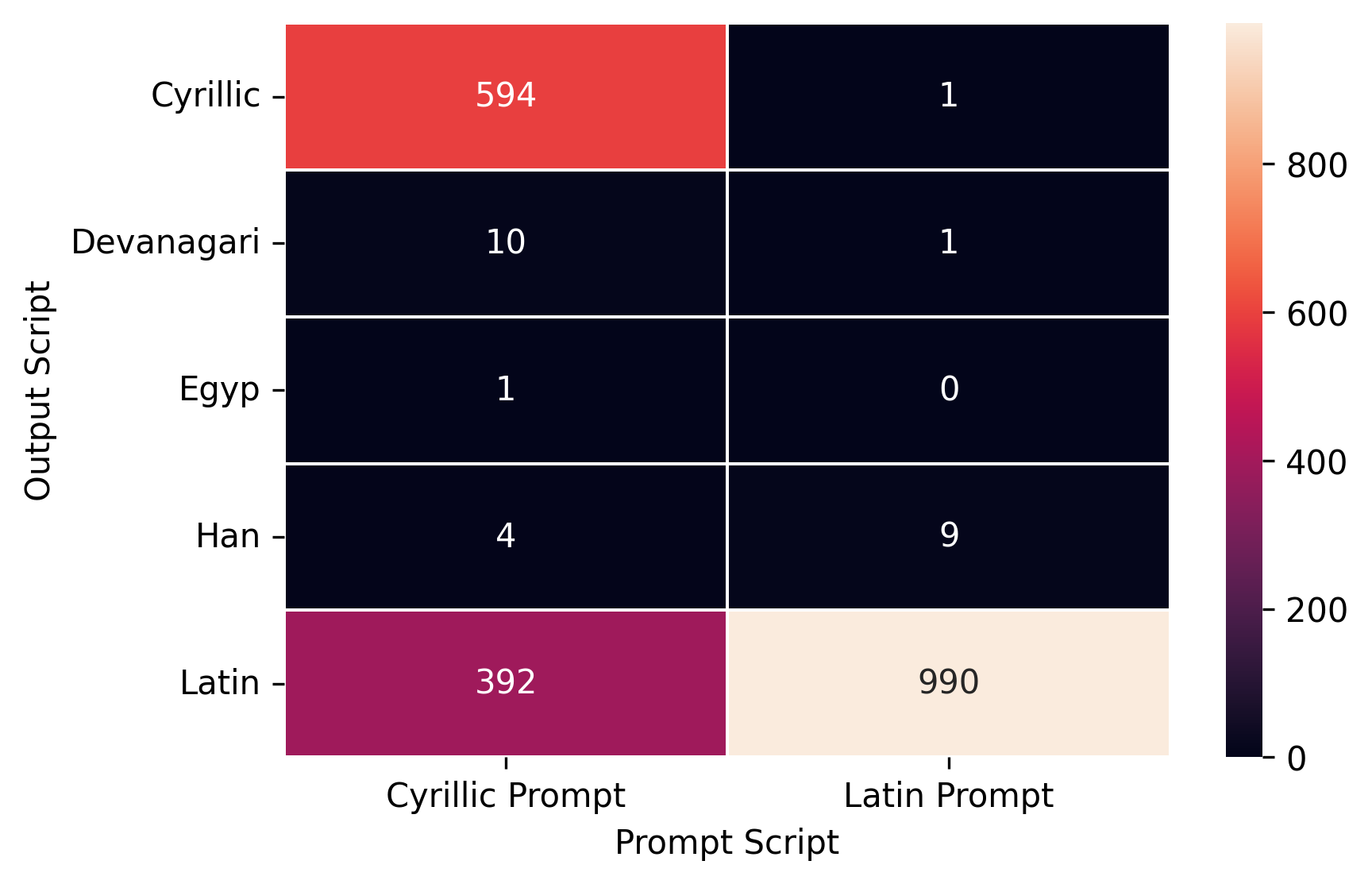}
        \caption{Mixtral-8x7B}
        \label{fig:pref_mixtral}
    \end{subfigure}
    \hfill
    \begin{subfigure}{0.45\textwidth}
        \centering
        \includegraphics[width=\linewidth]{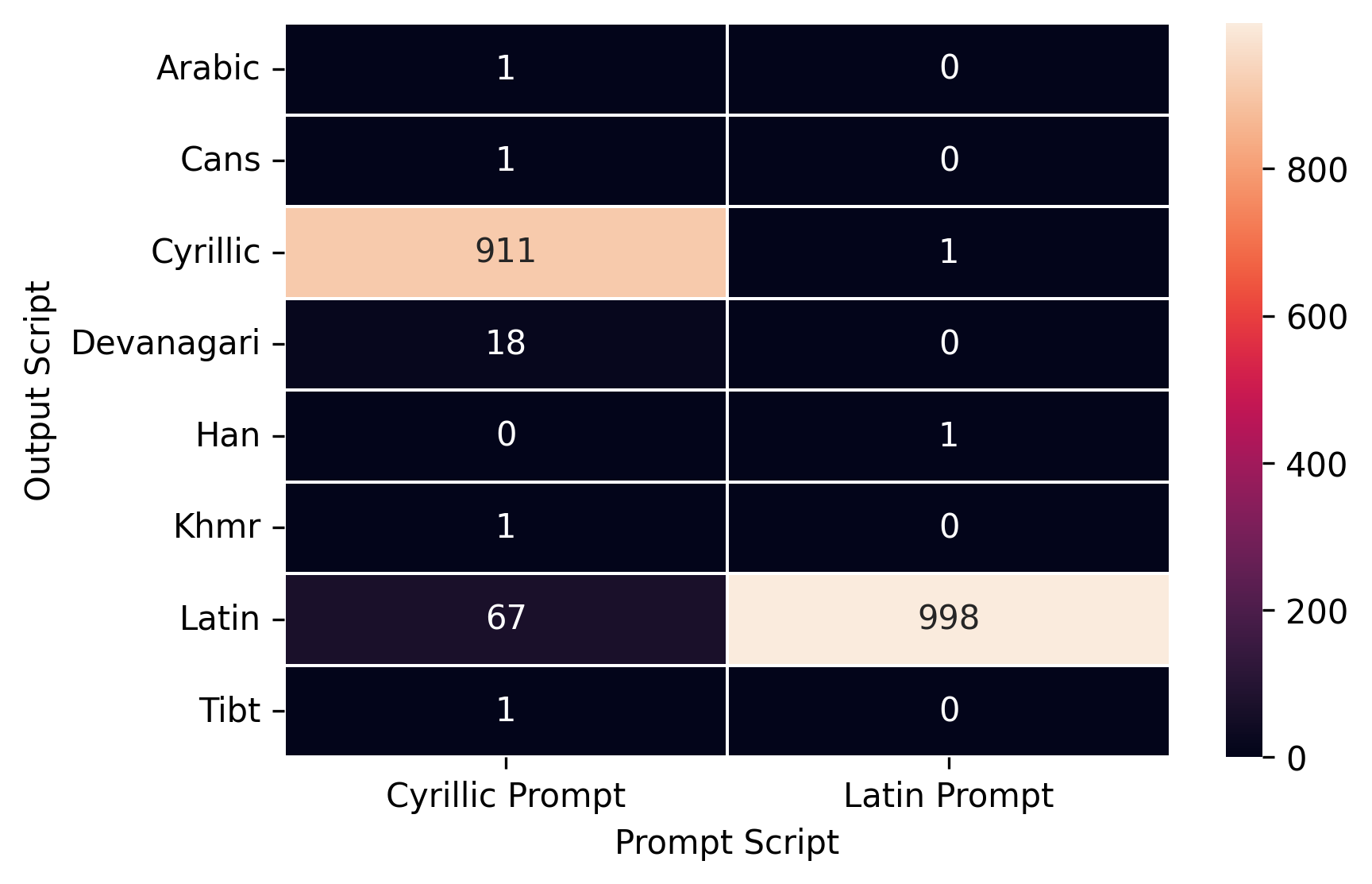}
        \caption{Mistral-7B}
        \label{fig:pref_mistral}
    \end{subfigure}

  \vspace{0.5em}

    \begin{subfigure}{0.45\textwidth}
        \centering
        \includegraphics[width=\linewidth]{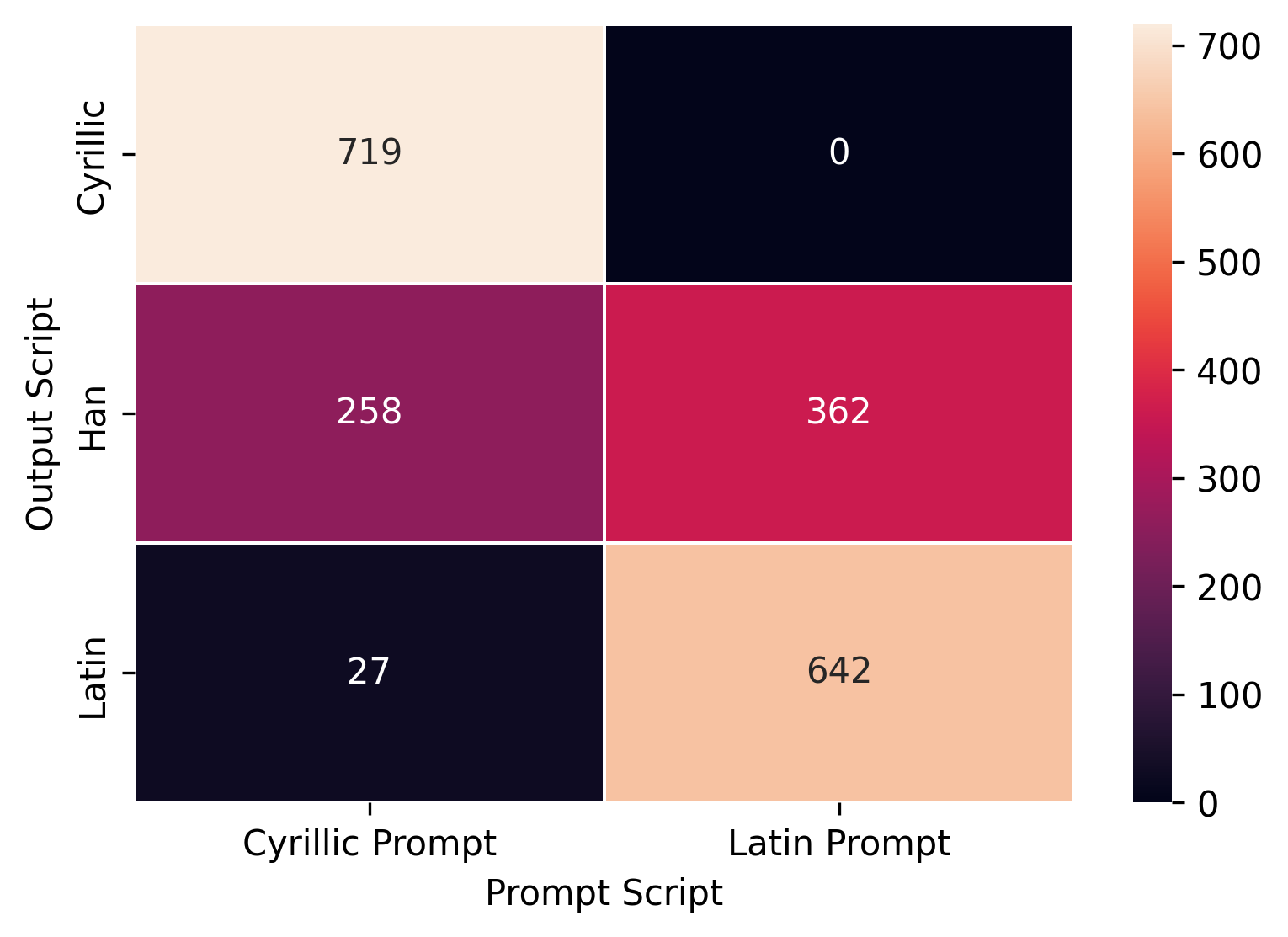}
        \caption{Qwen1.5-7B}
        \label{fig:pref_qwen}
    \end{subfigure}
    \hfill
    \begin{subfigure}{0.45\textwidth}
        \centering
        \includegraphics[width=\linewidth]{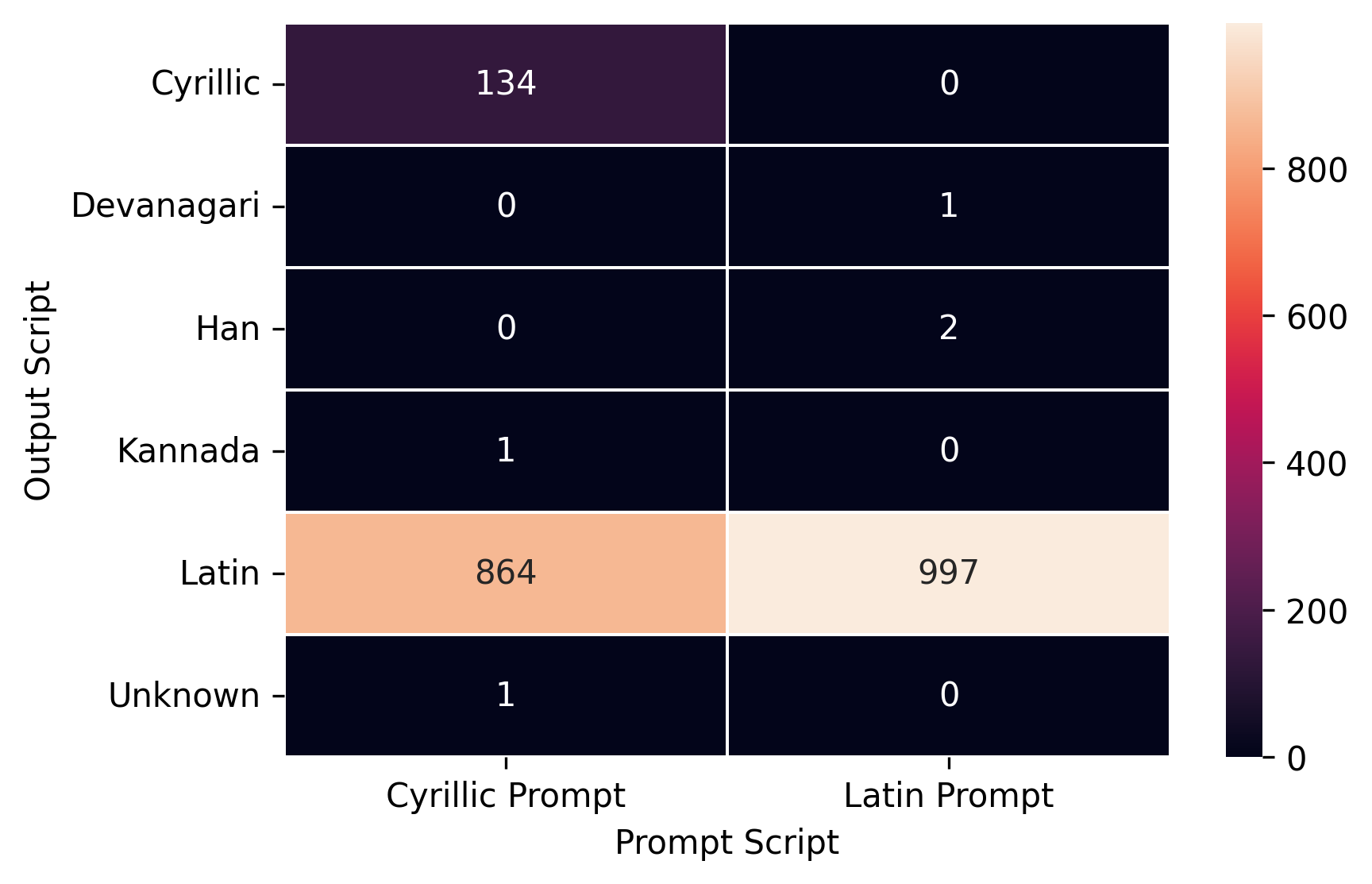}
        \caption{Llama-2-7B}
        \label{fig:pref_llama}
    \end{subfigure}

     \vspace{0.5em}
    \begin{subfigure}{0.45\textwidth}
        \centering
        \includegraphics[width=\linewidth]{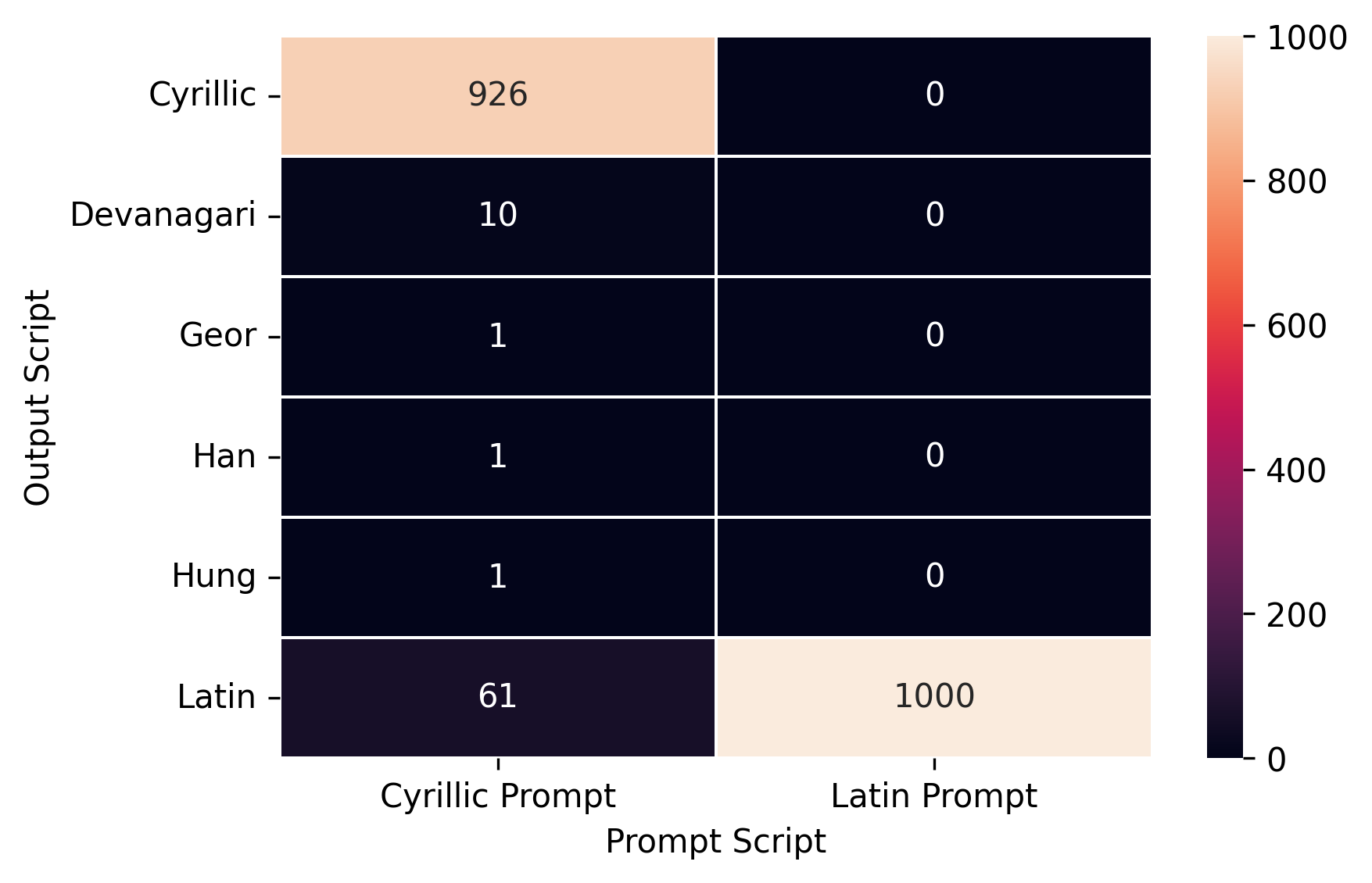}
        \caption{Deepseek-Distill-Llama-8B}
        \label{fig:pref_ds_llama}
    \end{subfigure}

    \caption{Distribution of output scripts for Latin (right column) and Cyrillic (left column) input prompts for open-source SLMs. Lighter colors indicate higher frequency}
    \label{fig:detail_preferences_appendix_SLMs}
\end{figure*}

\begin{figure*}[!h]
    \centering
    \begin{subfigure}{0.43\textwidth}
        \centering
        \includegraphics[width=\linewidth]{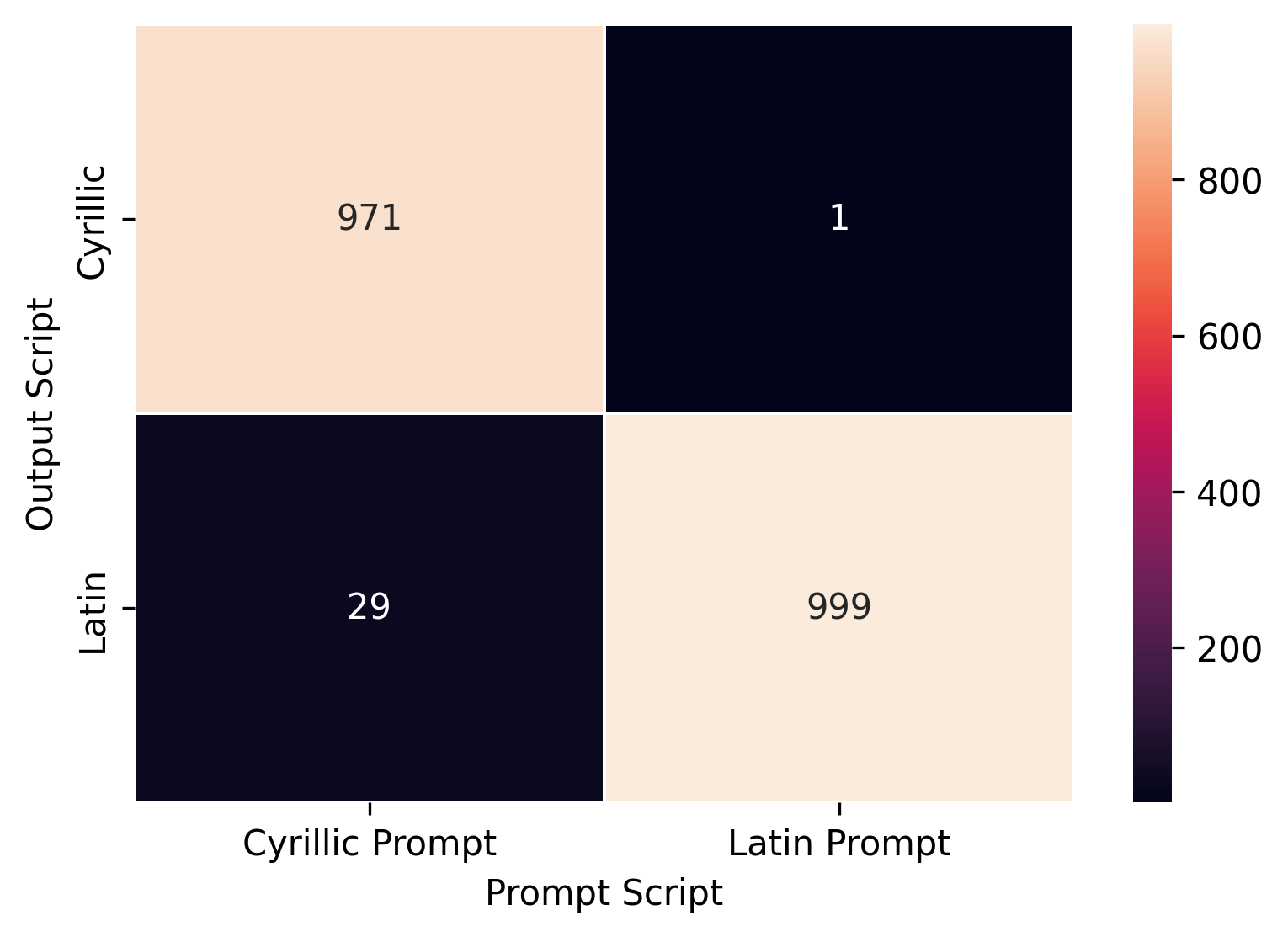}
        \caption{gpt-4o-mini}
        \label{fig:pref_4o-mini}
    \end{subfigure}
    \hfill
    \begin{subfigure}{0.43\textwidth}
        \centering
        \includegraphics[width=\linewidth]{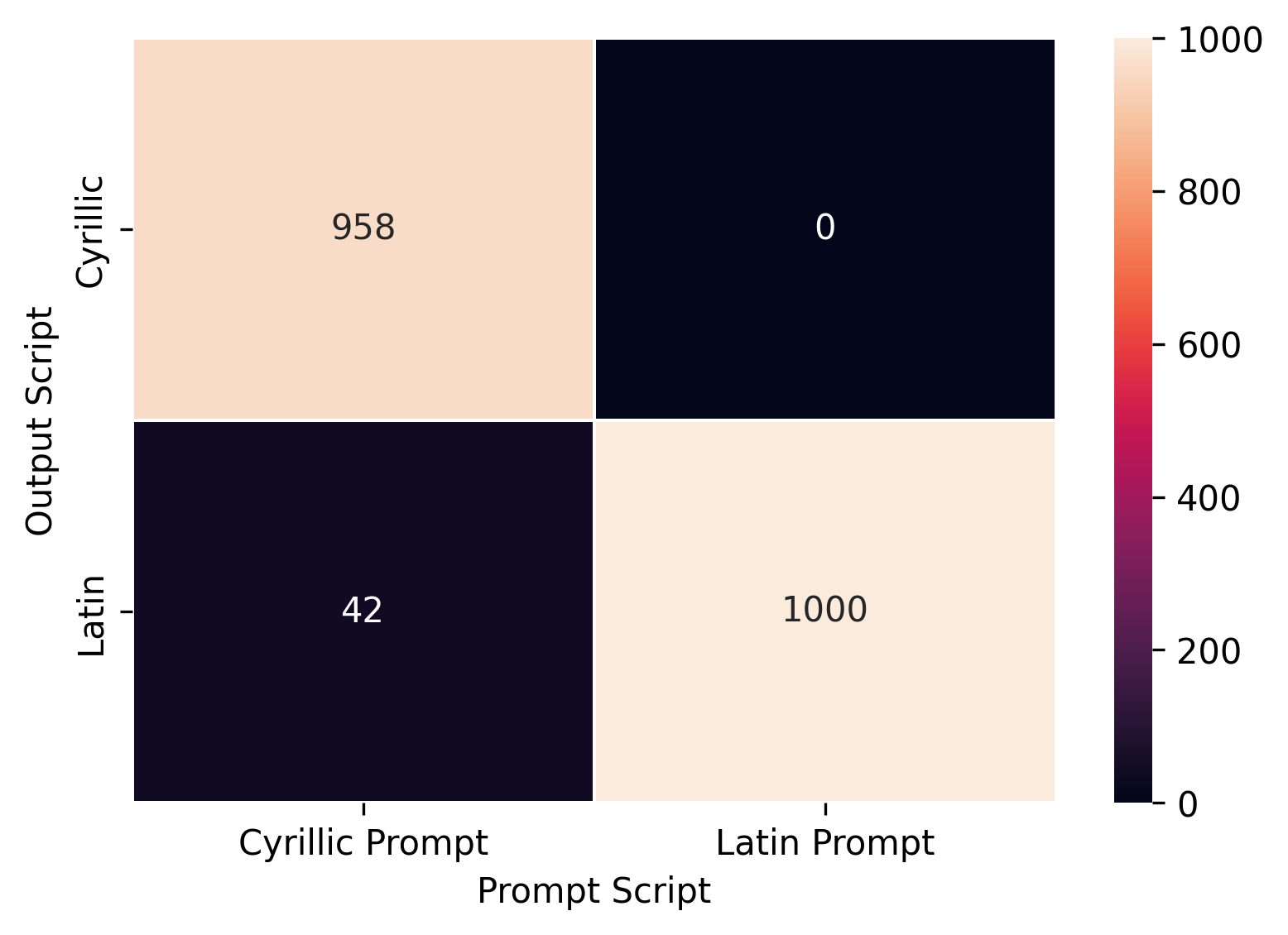}
        \caption{gpt-4o}
        \label{fig:pref_4o}
    \end{subfigure}

  \vspace{0.5em}

    \begin{subfigure}{0.43\textwidth}
        \centering
        \includegraphics[width=\linewidth]{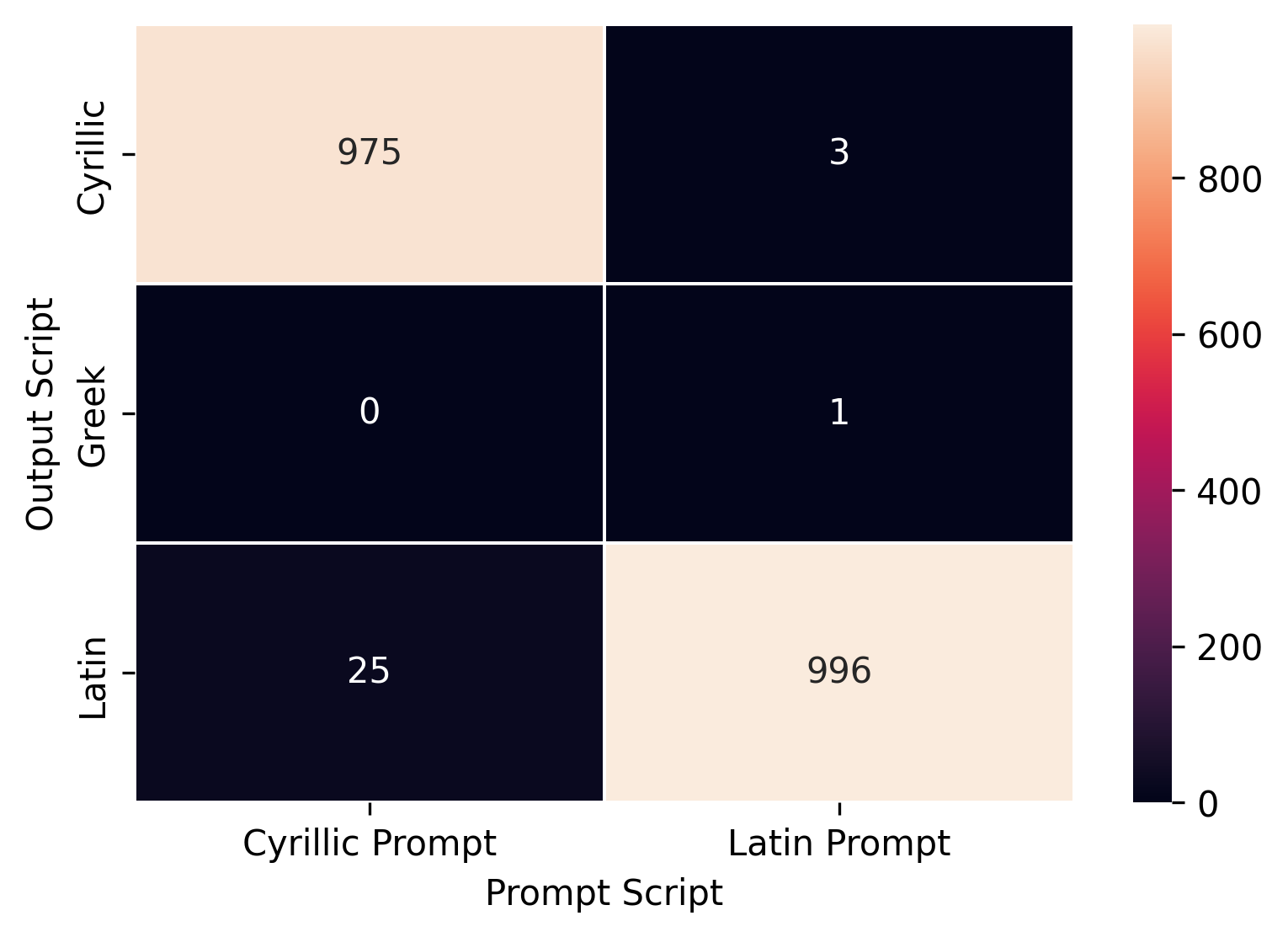}
        \caption{Gemini-2.5-flash}
        \label{fig:pref_flash}
    \end{subfigure}
     \hfill
    \begin{subfigure}{0.43\textwidth}
        \centering
        \includegraphics[width=\linewidth]{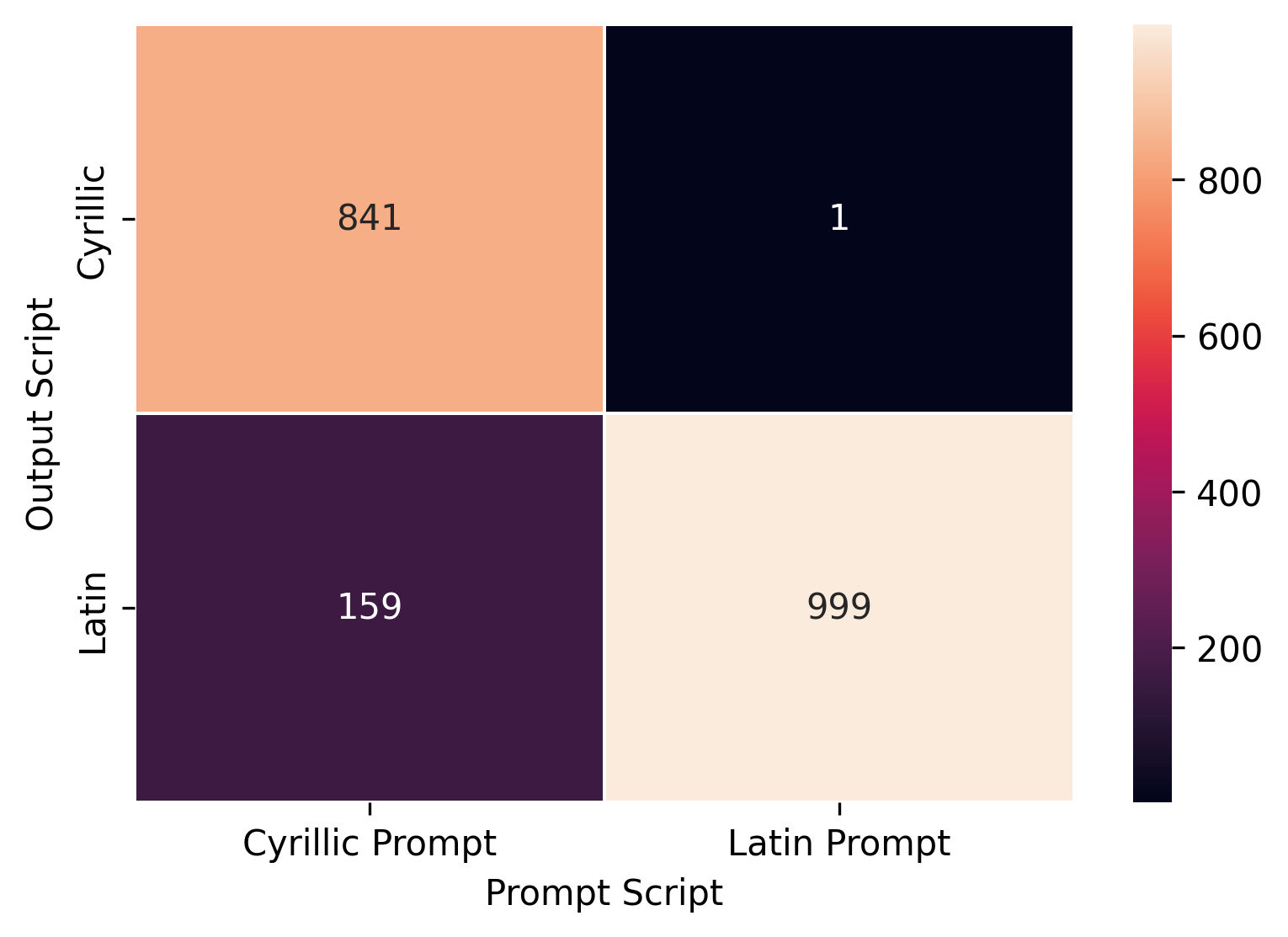}
        \caption{Gemini-2.5-flash-lite}
        \label{fig:pref_flash-lite}
    \end{subfigure}

    \vspace{0.5em}

    \begin{subfigure}{0.43\textwidth}
        \centering
        \includegraphics[width=\linewidth]{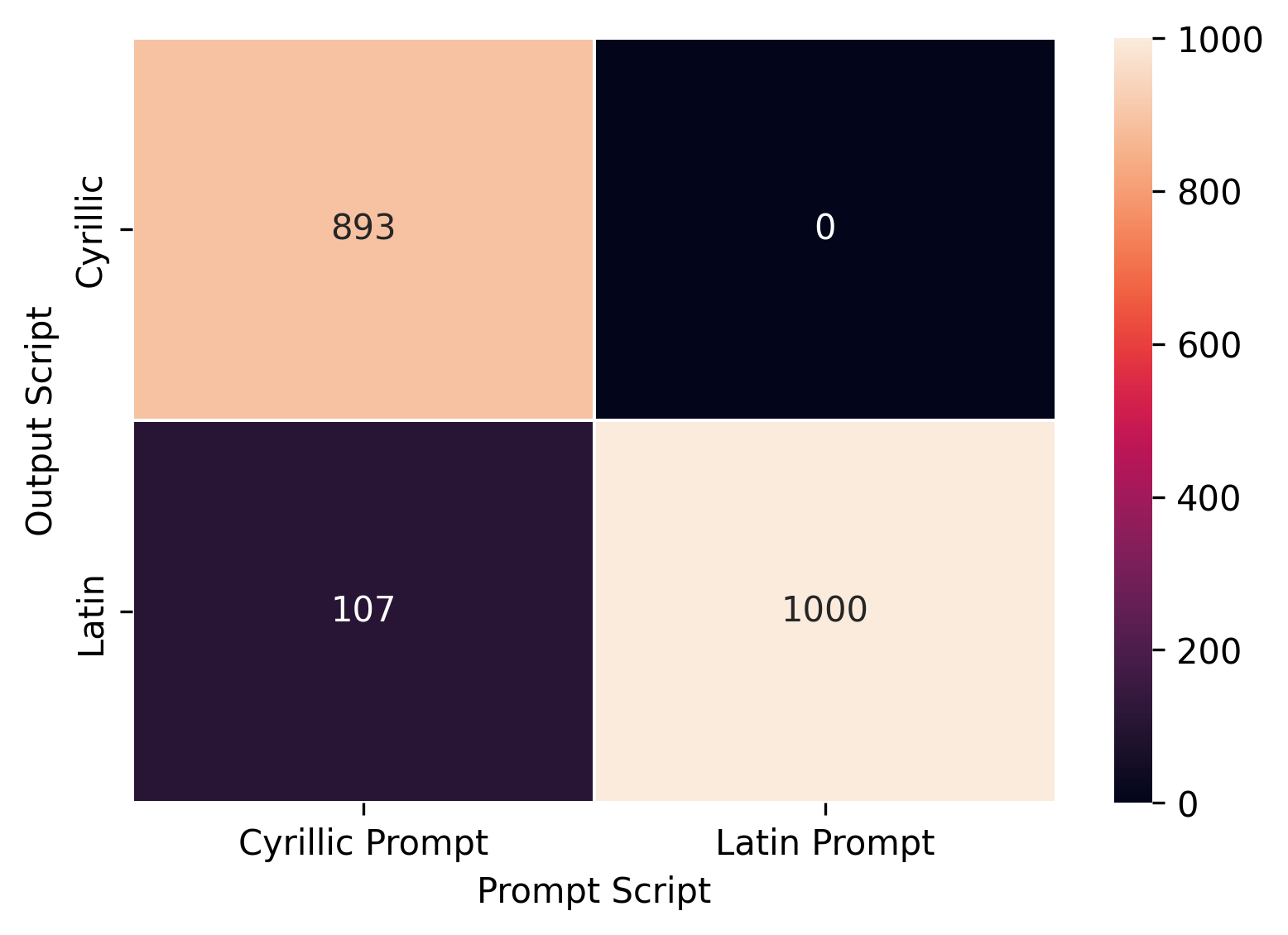}
        \caption{claude-3-5-haiku}
        \label{fig:pref_haiku}
    \end{subfigure}
    \hfill
     \begin{subfigure}{0.43\textwidth}
        \centering
        \includegraphics[width=\linewidth]{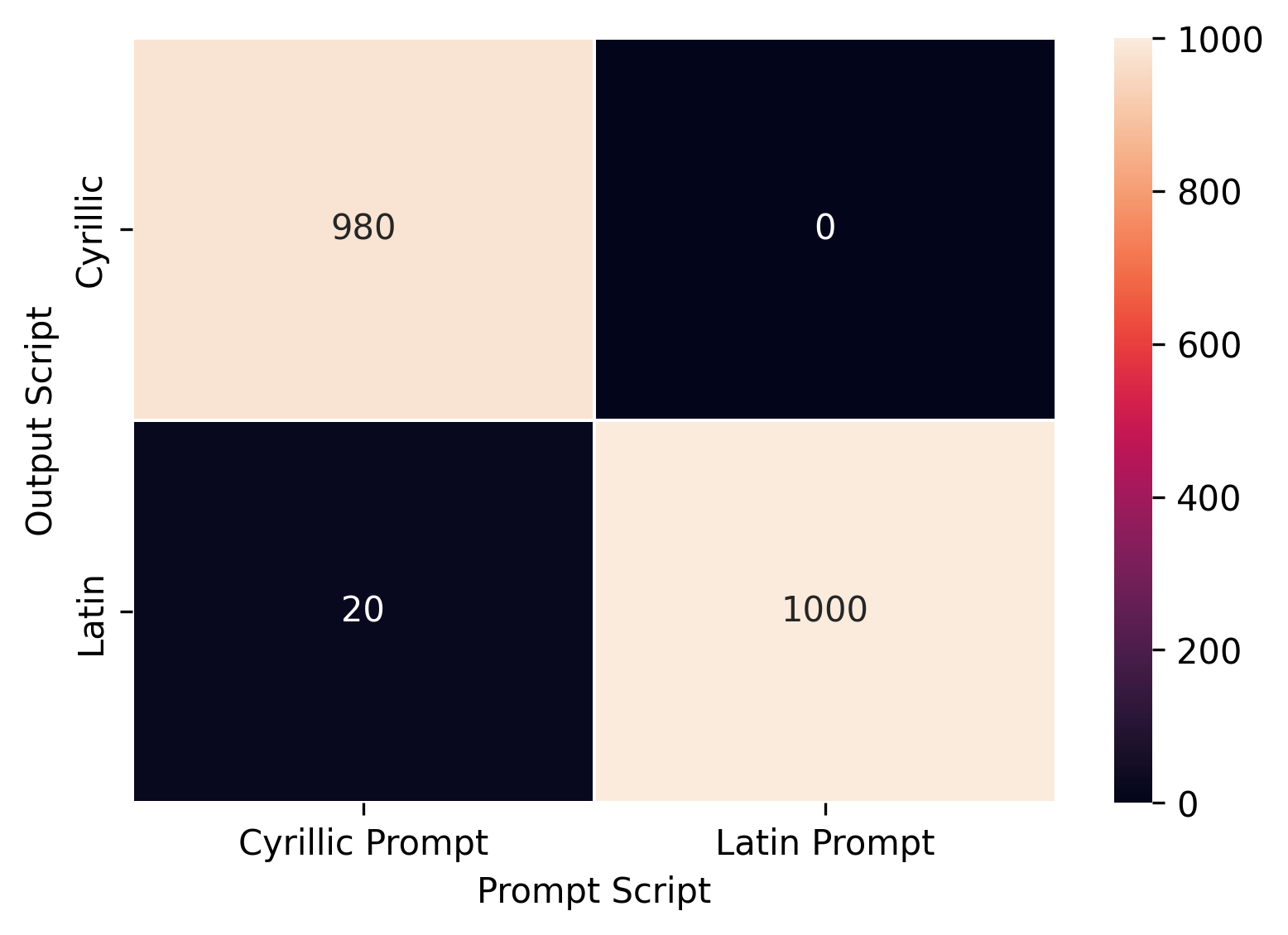}
        \caption{Deepseek-chat}
        \label{fig:pref_DS-chat}
    \end{subfigure}

    \vspace{0.5em}

    \begin{subfigure}{0.43\textwidth}
        \centering
        \includegraphics[width=\linewidth]{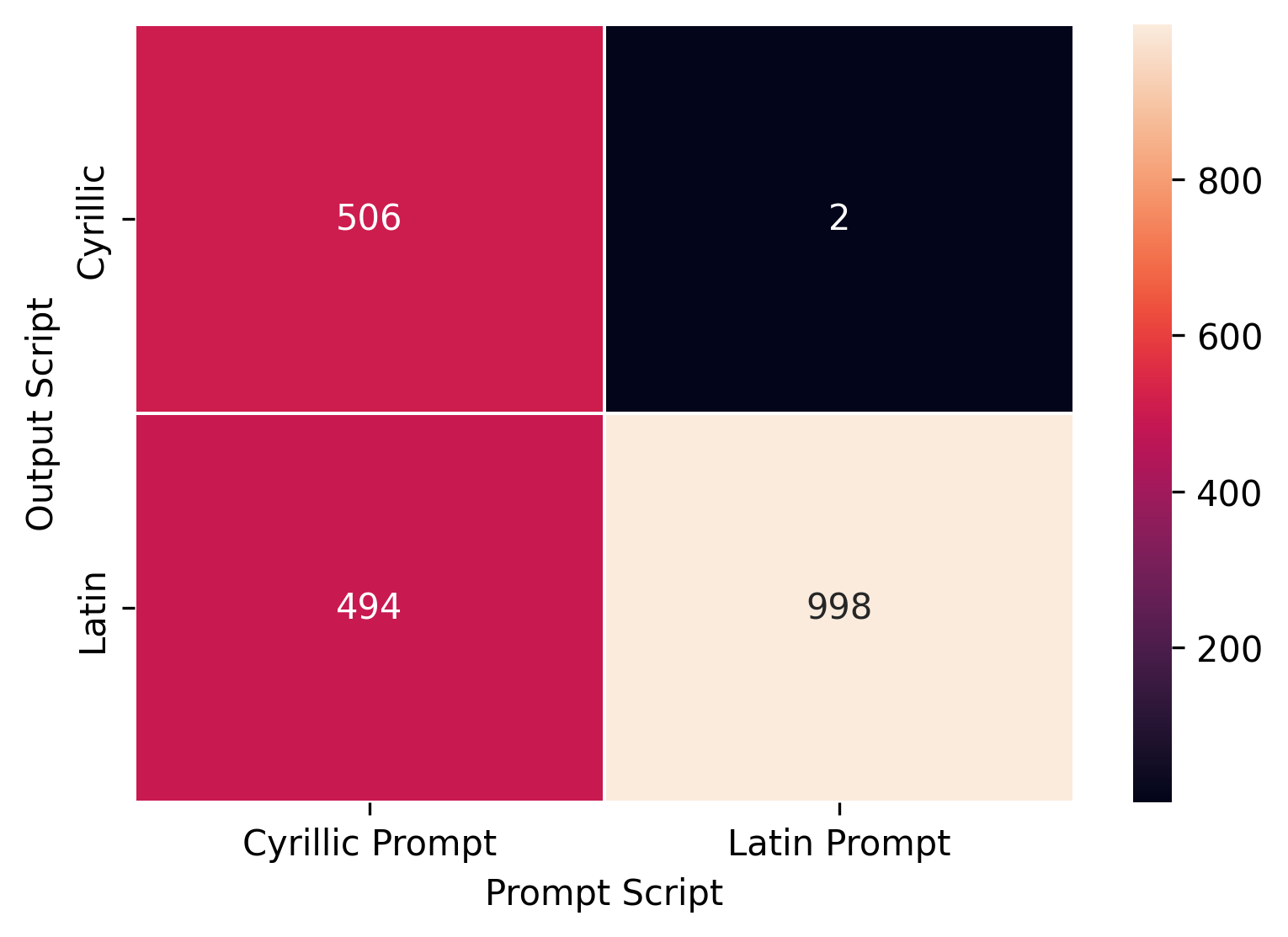}
        \caption{Claude-Sonnet-4}
        \label{fig:pref_sonnet}
    \end{subfigure}

    \caption{Distribution of output scripts for Latin (right column) and Cyrillic (left column) input prompts for proprietary LLMs. Lighter colors indicate higher frequency}
    \label{fig:detail_preferences_appendix_LLMs}
\end{figure*}

Details not included in Section~\ref{s:results} of scripts used by the models when they are unfaithful to the input script are given in Figures~\ref{fig:detail_preferences_appendix_SLMs} (open-source SLMs) and \ref{fig:detail_preferences_appendix_LLMs} (proprietary LLMs).

\section{Prompts}\label{a:prompts}

The five prompts we used for the script forcing are given in the prompts 1.1 to 1.5.

\begin{figure*}[!ht]
    \centering
    \includegraphics[width=\linewidth]{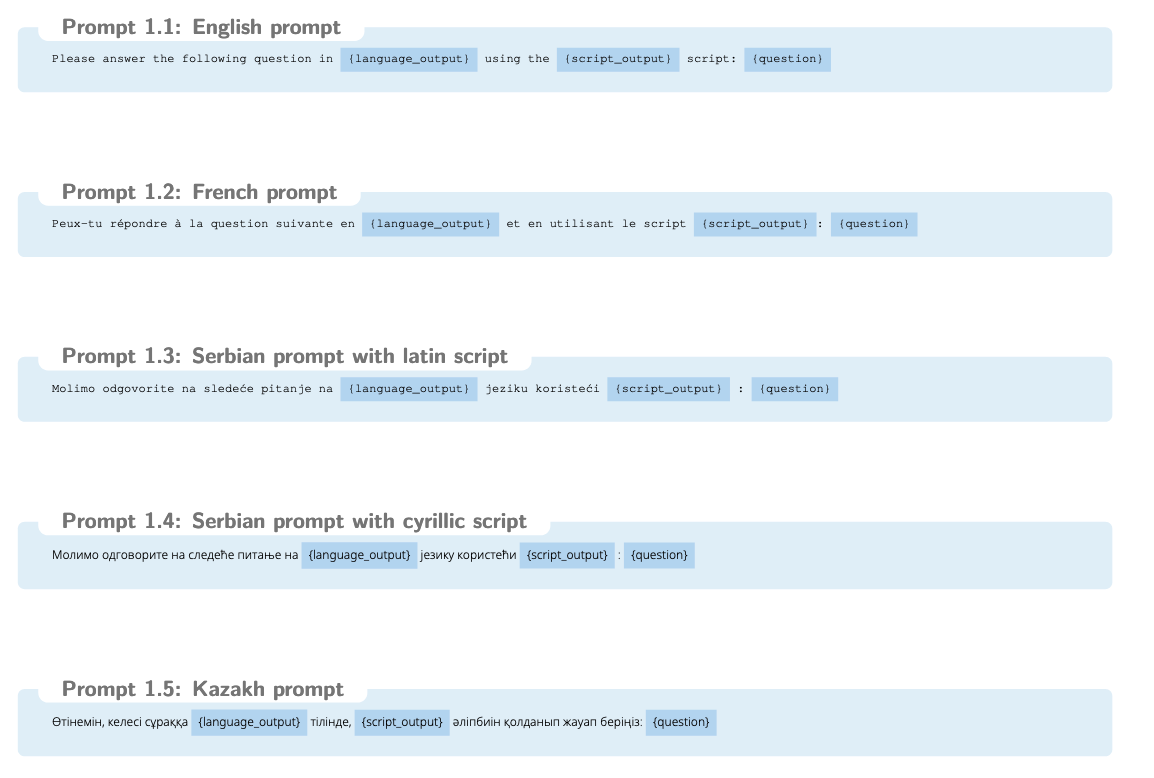}
    \label{fig:prompts_forcing}
\end{figure*}

\section{Script forcing configurations}\label{app:forcing_config}

The detail of the configurations used for the script are given in Table~\ref{tab:config_script_forcing}.

\newcounter{nbExp}
\stepcounter{nbExp}

\begin{table*}[!h]
\centering
\label{tab:example}
\begin{tabularx}{\textwidth}{XXXXXX}
\toprule

\textbf{Nb} & \textbf{Language prompt} & \textbf{Script prompt} & \textbf{Language Question } & \textbf{Requested language output } & \textbf{Requested script output} \\

\midrule
\thenbExp \stepcounter{nbExp} & French & Latin & French & French & Latin \\

\thenbExp \stepcounter{nbExp}  & French & Latin & French & French & Cyrillic  \\

\thenbExp \stepcounter{nbExp}  & English & Latin & French & French & Latin  \\

\midrule

\thenbExp \stepcounter{nbExp}  & English & Latin & Kazakh & Kazakh & Cyrillic \\

\thenbExp \stepcounter{nbExp}  & English & Latin & Kazakh & Kazakh &  Latin  \\

\thenbExp \stepcounter{nbExp}  & English & Latin & Kazakh & Kazakh & Arabic \\

 \thenbExp \stepcounter{nbExp}  & Kazakh & Cyrillic & Kazakh & Kazakh & Cyrillic \\

\thenbExp \stepcounter{nbExp}  & Kazakh & Cyrillic & Kazakh & Kazakh &  Latin \\

\thenbExp \stepcounter{nbExp}  & Kazakh & Cyrillic & Kazakh & Kazakh & Arabic \\

\midrule

\thenbExp \stepcounter{nbExp}  & English & Latin & Serbian & Serbian & Cyrillic \\

\thenbExp \stepcounter{nbExp}  & English & Latin & Serbian & Serbian &  Latin \\

\thenbExp \stepcounter{nbExp}  & Serbian & Latin & Serbian & Serbian & Cyrillic \\

\thenbExp \stepcounter{nbExp}  & Serbian & Latin & Serbian & Serbian &  Latin \\

\thenbExp \stepcounter{nbExp}  & Serbian & Cyrillic & Serbian & Serbian & Cyrillic \\

\thenbExp \stepcounter{nbExp}  & Serbian & Cyrillic & Serbian & Serbian &  Latin \\

\bottomrule
\end{tabularx}
\caption{Detail of configurations for the script forcing experiment (see Section~\ref{s:setup}).}
\label{tab:config_script_forcing}
\end{table*}

\section{Script Identification}\label{app:script_id}

Following \citet{holtermann-etal-2024-evaluating}, we identify output scripts using \texttt{glotLID}'s \textit{model\_v3}.\footnote{\url{https://huggingface.co/cis-lmu/glotlid}}
This model outputs a ranked list of scripts detected in the text along with their respective coverage percentages.
Since generated outputs may contain script-agnostic content such as mathematical formulas, we classify each text according to its script with the highest coverage.

\paragraph{Script identification without GlotLID}
To assess how much our findings depend on GlotLID, we also measured script compliance with a second method. For each output, we count the letters in each script, ignoring digits, punctuation and whitespace. We then compute the proportion of letters written in each script and count an output as in the requested script if the proportion of the requested is at least 0.8.

Across all 180 pairs, the two methods agree on 93.9\% of outputs. Agreement is higher for API models (96.7\%) than open models (89.9\%).
Only 13 of the 180 pairs fall below 80\% agreement, 10 of which involve open models.

We conducted an error analyses that reveals 3 types of errors.
1. Answers written in the requested script sometimes quote original names or titles in another script. 2. Some outputs combine two scripts, for instance when a model adds a short English sentence before the answer ("Here's the response in Kazakh using the Arabic script:"). GlotLID then assigns a single label, usually the script of the longest segment. 3. GlotLID sometimes misclassifies very short outputs in another script.

\section{Qwen results}\label{app:qwen}

Qwen1.5-7B exhibits markedly different behavior from all other tested models, frequently generating text in unexpected writing systems--most notably, in scripts that are unnatural for the intended language. This observation aligns with prior reports \citet{mitrovic-etal-2026-llms} that Qwen models sometimes produce Chinese text in unrelated contexts.

\paragraph{Script adaptation}
Qwen1.5 is the only model showing poor fidelity to Latin inputs (63.9\%). However, this cannot be attributed to an inverse preference for Cyrillic, as its fidelity in that case is similarly low (71.6\%). Rather, Qwen demonstrates a general lack of consistency across both writing systems. This broad unreliability in script selection appears to stem from the frequent generation of Chinese outputs, even when the expected script is neither Chinese nor ideographic.

\paragraph{Script forcing}
Across all configurations, Qwen1.5 yields consistently lower scores than other models, particularly for natural (language–script) pairs. For instance, when prompted in French with the Latin script, its fidelity scores are only 50.9\% and 13.2\% for French and English prompts, respectively--while the lowest score among all other models for this same configuration is 99.9\%.
Moreover, its highest score across all tested configurations does not exceed 67.9\%.

As in the adaptation setting, Qwen's generations are frequently contaminated by the unexpected insertion of Chinese text. Because the Hani script is never part of the expected configurations, these outputs systematically reduce its fidelity scores. This behavior indicates that Qwen1.5's issues extend beyond occasional noise, revealing a broader instability in how the model controls and maintains the intended script.

\section{Technical Configurations}\label{app:tech_details}

All experiments involving small language models (SLMs) were conducted on a server equipped with an NVIDIA L40S GPU.
Both SLMs and large language models (LLMs) were prompted using the \texttt{LangChain} framework, with the temperature parameter set to 0 for all runs.

The SLMs were downloaded via the HuggingFace interface \citep{wolf2020huggingfacestransformersstateoftheartnatural}. Each SLM was executed in its default configuration without any additional fine-tuning or optimization. All models were evaluated sequentially to ensure reproducibility and hardware consistency across runs.

\subsection{Running Time}

Approximate execution times (per 1,000 prompts) are reported below in Table~\ref{tab:exec_time}.

\begin{table}[!h]
\centering
\begin{tabular}{lcc}
\toprule
Model Type & Adaptation & Forcing \\
\midrule
Mixtral (Quantized) & $\sim$15 hrs & $\sim$5 hrs \\
Other SLMs & $\sim$7 hrs & $\sim$2 hrs \\
LLMs (via API) & $\sim$1-5 hr & $\leq$1-3 hrs \\
\bottomrule
\end{tabular}
\caption{Approximate execution times per 1,000 prompts}
\label{tab:exec_time}
\end{table}

\subsection{Costs}

Across all experiments, the total API costs were approximately
\$55, with the majority attributed to Claude-Sonnet-4 and GPT-4o.

All reported costs reflect combined input and output token.
\begin{table}[!h]
\centering
\begin{tabular}{lcc}
\toprule
Model & Adaptation & Forcing \\
\midrule
GPT-4o-mini & $\sim$\$0.5 & $\sim$\$1 \\
GPT-4o & $\sim$\$3 & $\sim$\$10 \\
DeepSeek-Chat & $\sim$\$0.5 & $\sim$\$2 \\
Claude-3.5-Haiku & $\sim$\$1.5 & $\sim$\$5 \\
Claude-Sonnet-4 & $\sim$\$5 & $\sim$\$20 \\
Gemini-Flash & $\sim$\$1 & $\sim$\$3 \\
Gemini-Flash-Lite & $\leq$\$0.5 & $\sim$\$0.5 \\
\midrule
\textbf{Total per experiment} & $\sim$\$10 & $\sim$\$45 \\
\bottomrule
\end{tabular}
\caption{Approximate API costs for each experimental configuration.}
\label{tab:api_costs}
\end{table}

\end{document}